\documentclass[10pt,twocolumn,letterpaper]{article}
\usepackage[T1]{fontenc}
\usepackage[pagenumbers]{cvpr}
\usepackage{mathtools}
\usepackage{bm}
\usepackage{multirow}
\newsavebox{\tablebox}
\newcommand{\fittable}[2]{%
  \begingroup\small
  \setlength{\tabcolsep}{4pt}%
  \sbox{\tablebox}{#2}%
  \ifdim\wd\tablebox>#1\relax
    \resizebox{#1}{!}{\usebox{\tablebox}}%
  \else\usebox{\tablebox}\fi
  \endgroup
}

\usepackage{colortbl}
\definecolor{oursblue}{HTML}{E5EFFC}   
\definecolor{ourspurple}{HTML}{EFE7F6} 

\newcommand{\method}{DepthMSA}
\newcommand{\dsmosi}{CMU-MOSI}
\newcommand{\dsmosei}{CMU-MOSEI}
\newcommand{\dssims}{CH-SIMS}
\newcommand{\up}{$\uparrow$}
\newcommand{\dn}{$\downarrow$}

\definecolor{cvprblue}{rgb}{0.21,0.49,0.74}
\usepackage[pagebackref,breaklinks,colorlinks,allcolors=cvprblue]{hyperref}
\title{DepthMSA: Combinatorial Training and Budget-Aware Depth Allocation for Multimodal Sentiment Analysis}
\author{Pan Wang$^{1*}$ \quad Jie Peng$^{4}$ \quad Zhimin Li$^{1}$ \quad Jingxian Chen$^{1}$ \quad Hui Ji$^{1}$\\
Huanrui Yang$^{3}$ \quad Pengfei Zhou$^{1}$ \quad Tianlong Chen$^{2}$ \quad Jingtong Hu$^{1}$\\[0.4em]
$^{1}$University of Pittsburgh \quad $^{2}$University of North Carolina at Chapel Hill\\
$^{3}$University of Arizona \quad $^{4}$Foci Labs
}
\begin{document}

\twocolumn[{%
\renewcommand\twocolumn[1][]{#1}%
\maketitle
\vspace{-1.6em}
\begin{center}
  \captionsetup{type=figure}
  \includegraphics[width=0.88\textwidth]{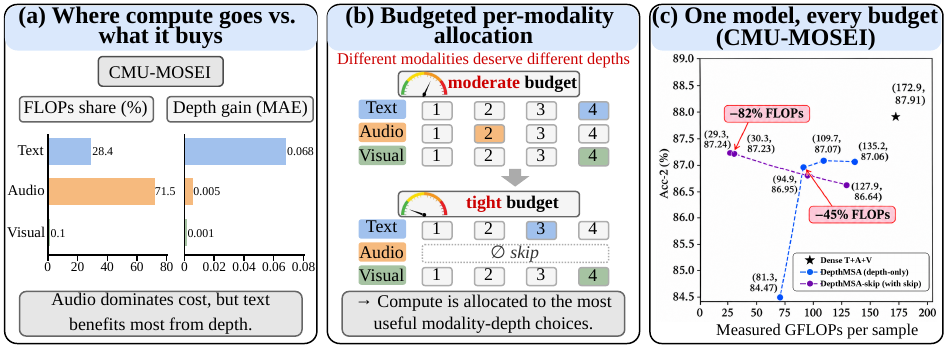}
  \captionof{figure}{\textbf{Do all modalities need full-depth encoding for multimodal sentiment analysis?} (a) On \dsmosei{}, audio takes 71.5\% of the encoder FLOPs, but truncating it to a quarter of its depth changes the error by only 0.005 MAE, against 0.068 for text. (b) \method{} gives each modality its own exit under a compute budget (illustrative configurations; skipping a modality requires the skip-enabled model). (c) One trained model serves every budget: the depth-only model stays within one point of the dense model (star) with 45\% fewer measured FLOPs, and the separately trained skip-enabled model with 82\% fewer. Mean of 3 seeds.}
  \label{fig:teaser}
\end{center}
\vspace{0.4em}
}]
\begingroup
\renewcommand\thefootnote{}\footnotetext{$^{*}$ Corresponding author: {\tt pan.wang@pitt.edu}}
\endgroup

\begin{abstract}
State-of-the-art multimodal sentiment analysis (MSA) models rely on large pretrained encoders, yet typically process every modality to its full depth, regardless of its computational cost or predictive contribution. This raises a fundamental question: how should limited computation be allocated across modalities to preserve predictive performance? We reveal a substantial mismatch between the computational cost of modality encoders and the predictive benefit of processing them more deeply, suggesting that different modalities require different encoding depths. However, enabling heterogeneous depths introduces a key challenge: representations extracted at different depths must still be effectively fused. To address this, we propose \method{}, a budget-aware framework that supports flexible modality depths within a single MSA model. Specifically, we introduce combinatorial anytime training, which supervises the fusion head across all cross-modal depth combinations using a single full-depth encoder pass. At inference, a training-free, cost-aware allocator selects the configuration with the lowest validation-calibrated error under a sample-specific FLOPs budget. On \dsmosei{}, \method{} maintains accuracy within one percentage point of the full-depth model while reducing measured FLOPs by 45\%, or by 82\% with a separately trained skip-enabled variant. When a CLIP visual encoder dominates computation, the allocation shifts toward reducing visual depth, achieving 56\% fewer FLOPs at unchanged accuracy. Compared with ADMN, an adaptive layer-allocation method that trains a controller for each budget, the skip-enabled variant reaches its accuracy with 3.2$\times$ fewer FLOPs and lower latency, without any per-budget training. Our project is available at \url{https://pwang322.github.io/DepthMSA/}.
\end{abstract}

\section{Introduction}
\label{sec:intro}

Multimodal sentiment analysis (MSA) aims to understand human sentiment by integrating linguistic, acoustic, and visual cues~\cite{mosi,mosei,chsims,pmlr-v306-wang26cz,wang2020makes}. Recent advances increasingly rely on large pretrained encoders, such as RoBERTa~\cite{roberta}, data2vec~\cite{data2vec}, HuBERT~\cite{hubert}, and CLIP~\cite{radford2021learning}, to extract rich modality-specific representations before fusion~\cite{mmml}. Despite their different computational costs and predictive roles, these encoders are typically executed to full depth for every input, leading to substantial inference overhead. This raises a fundamental question: \emph{how should a limited computational budget be allocated across modalities to best preserve sentiment prediction accuracy?}

Existing approaches improve efficiency by reducing model size, removing modalities, or dynamically adjusting network depth~\cite{deebert,pabee,berxit,mue,admn,wang2026models}. However, smaller models and fixed modality subsets are tied to particular computational settings, while conventional early-exit methods do not explicitly address heterogeneous depths across interacting modalities. ADMN~\cite{admn} allocates layers across modalities, but relies on layer-count budgets and a learned controller for each budget rather than on input-dependent FLOPs. Moreover, independently truncating modality encoders does not ensure that their intermediate representations can be effectively fused. This motivates a single MSA model that supports flexible modality depths under varying compute budgets without retraining.

We revisit the relationship between computational cost and the predictive benefit of encoding depth. Surprisingly, the most expensive modality is not necessarily the one that benefits most from deeper processing (\cref{fig:teaser}). On \dsmosei{}, audio consumes 71.5\% of encoder FLOPs, yet reducing its depth to one quarter increases the validation error by only 0.005 MAE, whereas the same truncation of text, which accounts for just 28.4\% of FLOPs, costs 0.068. Similar patterns emerge on \dsmosi{} and \dssims{}, while the computational bottleneck shifts to vision when using a CLIP encoder. These observations reveal a key insight: \emph{the importance of a modality does not necessarily reflect the benefit of encoding it more deeply}. Efficient MSA should therefore allocate computation according to depth-dependent predictive utility rather than uniformly processing every modality to full depth.

Motivated by this insight, we propose \method{} (\cref{fig:overview}), a framework that enables heterogeneous modality depths under varying compute budgets. First, \emph{combinatorial anytime training} equips each modality encoder with shared-weight exits and jointly supervises all cross-modal depth combinations. Since one full-depth encoder pass yields the features of all exits, the fusion head learns to fuse features from different depths without a separate pass for each configuration. Second, \emph{cost-aware depth allocation} calibrates the prediction error of each combination on validation data and selects the lowest-error configuration satisfying each sample's FLOPs budget. The allocator requires no training, so one model serves all budgets.

Experiments on \dsmosi{}, \dsmosei{}, and \dssims{} demonstrate substantial efficiency gains with competitive prediction performance. On \dsmosei{}, \method{} reduces measured FLOPs by 45\% with less than one percentage point of accuracy loss, while a separately trained skip-enabled variant achieves an 82\% reduction. With a CLIP visual encoder, the allocation shifts toward reducing visual depth, saving 56\% FLOPs at unchanged accuracy. Against ADMN~\cite{admn} on the same backbones, the skip-enabled variant matches its accuracy with 3.2$\times$ fewer FLOPs and 36\% lower latency, from one model trained once. Our contributions are:
\begin{itemize}[leftmargin=*,nosep]
    \item \textbf{Compute--utility insight.}
    We reveal a substantial mismatch between the computational cost of modality encoders and the predictive benefit of deeper encoding, motivating heterogeneous depth allocation in MSA.

    \item \textbf{Combinatorial anytime training.}
    We propose a framework DepthMSA, which jointly supervises cross-modal depth combinations to enable effective fusion across heterogeneous encoder depths, with optional modality skipping in a separately trained variant.

    \item \textbf{Cost-aware depth allocation.}
    We develop a validation-calibrated allocation strategy that selects modality-depth configurations based on their predictive utility and input-dependent FLOPs under per-sample compute budgets, without budget-specific retraining.
\end{itemize}

\section{Related Work}
\label{sec:related}

\noindent\textbf{Multimodal sentiment analysis.}
MSA has evolved from tensor-based fusion~\cite{tfn,lmf} and cross-modal attention~\cite{mult} to representation learning that disentangles modality-shared and specific information~\cite{misa}, introduces unimodal supervision~\cite{selfmm}, or improves cross-modal alignment and interaction~\cite{mmim,confede,almt,kuda,dlf,unimse}. More recently, MMML~\cite{mmml} leverages large pretrained encoders for text and audio, which we adopt as our backbone. While modality imbalance, particularly the dominance of text, has been widely studied~\cite{ogmge,wang2025dlf}, existing methods primarily seek to improve modality utilization and predictive accuracy. We instead investigate whether the computation spent on deeper modality encoding is justified by its incremental predictive benefit.

\noindent\textbf{Early exiting and dynamic networks.}
Early-exit methods~\cite{teerapittayanon2016branchynet,msdnet,deebert,fastbert,pabee,berxit} reduce inference cost by producing predictions at intermediate network depths. LayerDrop~\cite{layerdrop}, slimmable networks~\cite{yu2019slimmable}, and once-for-all networks~\cite{cai2020once} further support flexible computation within shared model weights. Non-uniform Transformers~\cite{vwt} also vary the compute allocation, but in network width rather than inference depth. In MSA, however, multiple encoders may exit at different depths, and their representations must still be fused effectively. Unlike independently trained exits or stochastic depth, our combinatorial anytime training explicitly supervises all cross-modal depth combinations, enabling flexible inference within a single model.

\noindent\textbf{Efficient multimodal inference.}
Existing approaches reduce multimodal computation through modality selection~\cite{panda2021adamml}, adaptive fusion~\cite{dynmm}, multimodal early exiting~\cite{mue,mmexit}, or visual token pruning~\cite{fastv,sparsevlm}. Most closely related, ADMN~\cite{admn} allocates a fixed layer budget across modalities according to input quality, using LayerDrop-trained backbones and budget-specific controllers. However, layer counts do not reflect the heterogeneous FLOPs of modality encoders, which also vary with input length. In contrast, \method{} computes the input-dependent FLOPs of each modality-depth configuration and selects the one with the lowest validation-calibrated error under a sample-specific FLOPs budget, without per-budget controllers.

\section{Method}
\label{sec:method}

\begin{figure*}[t]
  \centering
  \includegraphics[width=\textwidth]{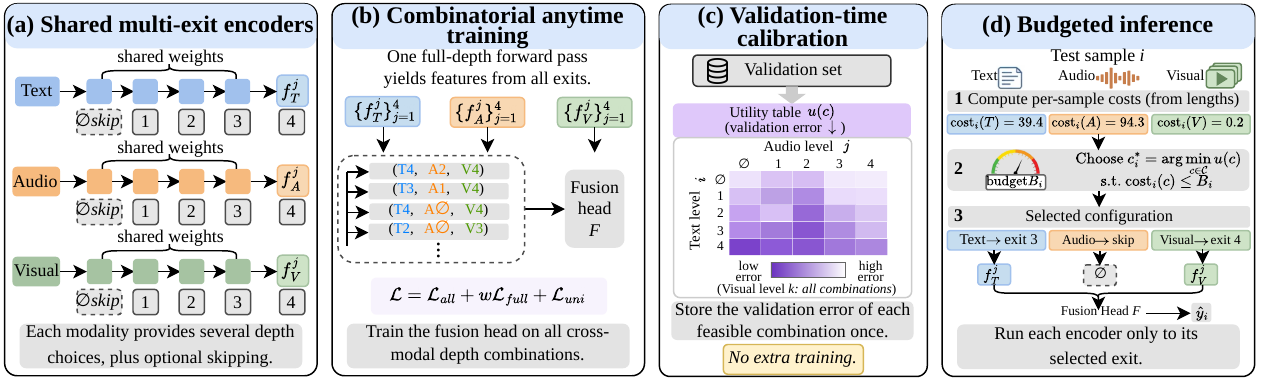}
  \caption{\textbf{Overview of \method{}.}
  (a) Each modality encoder provides $J=4$ shared-weight
  exits and an optional skip level.
  (b) Combinatorial anytime training supervises all
  cross-modal depth configurations using a single
  full-depth encoder pass.
  (c) Validation-time calibration estimates the prediction
  error of each configuration and constructs a utility table.
  (d) For each test sample, the allocator computes
  configuration costs, selects the lowest-error feasible
  configuration, and executes the corresponding encoder
  depths.}
  \label{fig:overview}
\end{figure*}

\subsection{Problem Formulation}
\label{sec:problem}

We consider an MSA model with $M$ modalities, where $M=3$
corresponds to text, audio, and vision. Given a sample
$\bm{x}_i=(x_{i,1},\ldots,x_{i,M})$ with sentiment label $y_i$,
each modality $m$ is processed by a pretrained Transformer
encoder $E_m$ with $L_m$ layers. A fusion head $F$ combines
the resulting modality features to predict the sentiment
score. We adopt the MMML architecture~\cite{mmml}, which
incorporates preceding utterances as context for text and
audio and uses an MLP for multimodal fusion.

Rather than executing every encoder to full depth, we
consider a \emph{depth configuration}
$\bm{c}=(c_1,\ldots,c_M)\in\mathcal{C}$,
where $c_m$ denotes the selected exit level of modality
$m$ and $\mathcal{C}$ is the set of supported configurations.
Each sample $i$ has a compute budget $B_i=\rho C_i$,
where $C_i$ is its full-depth modelled FLOPs and
$\rho\in(0,1]$ is the budget ratio.
We train a single model that supports
heterogeneous modality depths and selects, for each sample
and budget, the feasible configuration with the lowest
estimated error.

The budget constrains the modelled encoder computation,
as defined in \cref{sec:allocator}. Input-side computation
is included when modality skipping is supported, while
the lightweight readouts and fusion head are excluded.
We report measured end-to-end FLOPs separately in
\cref{sec:setup}.

\subsection{Framework Overview}
\label{sec:framework}

As shown in \cref{fig:overview}, \method{} builds modality
encoders with shared-weight exits (a), jointly trains their
cross-modal depth combinations (b), calibrates the predictive
utility of each configuration on validation data (c), and,
at inference, selects and executes a configuration within
the sample-specific compute budget (d).

\noindent\textbf{Shared-weight exits.}
As shown in \cref{fig:overview}(a), each modality encoder
provides $J$ exits at layers
$\ell_m^1<\cdots<\ell_m^J=L_m$.
We use $J=4$ evenly spaced exits.
Exit $j$ produces a feature
$f_{i,m}^{\,j}$ from the first $\ell_m^j$ encoder
layers through a lightweight readout.
Specifically, we use a pooler for text and masked
mean pooling followed by layer normalization for
audio and vision. All exits share the same encoder
weights, allowing deeper exits to reuse the computation
of shallower ones.

\noindent\textbf{Optional modality skipping.}
We additionally support a skip level $j=0$, which
replaces the encoded feature with a learned null vector
$f_{i,m}^{\,0}=\bm{z}_m$.
Unlike an intermediate exit, skipping bypasses the
entire modality encoder, including input-side computation
such as speech feature extraction or visual patch embedding.
We denote this separately trained variant by \method-skip.

Without skipping, each modality selects from
$\{1,\ldots,J\}$; with skipping, it selects from
$\{0,\ldots,J\}$. The resulting configuration space
contains $J^M=64$ or $(J+1)^M=125$ combinations,
respectively. For any configuration $\bm{c}$,
the prediction is
\begin{equation}
  \hat{y}_i(\bm{c})
  =F\left(
    f_{i,1}^{\,c_1},\ldots,f_{i,M}^{\,c_M}
  \right),
  \label{eq:prediction}
\end{equation}
where $\hat{y}_i(\bm{c})$ is the predicted sentiment
score of sample $i$ under configuration $\bm{c}$,
and $f_{i,m}^{\,c_m}$ is the feature of modality $m$
at its selected level $c_m$.

\subsection{Combinatorial Anytime Training}
\label{sec:training}

Figure~\ref{fig:overview}(b) illustrates how we train
the depth configurations constructed in (a).
Although each encoder provides multiple exits, their
representations may correspond to different semantic
levels. A fusion head trained only on full-depth features
is not explicitly optimized to combine, for example,
deep text representations with shallow audio features.
Moreover, independently supervising unimodal exits
does not account for their interactions during fusion.

We therefore introduce \emph{combinatorial anytime
training}, which jointly supervises the fusion head
across all cross-modal depth combinations.
Since the exits of each encoder share weights,
one full-depth forward pass provides features from
every exit. We then evaluate the lightweight fusion
head on their Cartesian product, as illustrated
in \cref{fig:overview}(b).

For a training sample $i$, the objective is
\begin{equation}
\begin{aligned}
  \mathcal{L}_i
  ={}&
  \underbrace{
    \frac{1}{|\mathcal{C}|}
    \sum_{\bm{c}\in\mathcal{C}}
    |\hat{y}_i(\bm{c})-y_i|
  }_{\mathcal{L}_{\mathrm{all},i}}\\
  &+w\underbrace{
    |\hat{y}_i(\bm{c}_{\mathrm{full}})-y_i|
  }_{\mathcal{L}_{\mathrm{full},i}}
  +\mathcal{L}_{\mathrm{uni},i},
\end{aligned}
\label{eq:loss}
\end{equation}
where $\mathcal{L}_{\mathrm{all},i}$ is the absolute
error averaged over all $|\mathcal{C}|$ supported
configurations, $\mathcal{L}_{\mathrm{full},i}$ is the
absolute error of the full-depth configuration
$\bm{c}_{\mathrm{full}}$, $w$ weights this full-depth
term ($w=1$ in our experiments), and
$\mathcal{L}_{\mathrm{uni},i}$ is the unimodal loss
defined in \cref{eq:uniloss}.
The first term jointly supervises all combinations,
enabling the fusion head to process heterogeneous-depth
features. The second emphasizes full-depth prediction
to mitigate performance degradation from joint training
across depths.
The third term keeps the base model's unimodal
loss at every exit of every modality:
\begin{equation}
  \mathcal{L}_{\mathrm{uni},i}
  =\sum_{m=1}^{M}\frac{1}{J}
  \sum_{j=1}^{J}
  |U_m(f_{i,m}^{\,j})-y_{i,m}|,
  \label{eq:uniloss}
\end{equation}
where $j$ indexes the exits, $U_m$ is the unimodal
regression head of modality $m$, and $y_{i,m}$ is the modality-specific sentiment
label when available, or the overall label $y_i$
otherwise. The skip level is excluded
from unimodal supervision.

We select the checkpoint minimizing
$\mathcal{L}_{\mathrm{all}}$, the mean of
$\mathcal{L}_{\mathrm{all},i}$ over the validation set, favoring performance
across configurations rather than at a single depth.
The encoders are evaluated only once per training
step; the additional computation comes from the
exit readouts and fusion evaluations. In our
measurements, the training time per step stays
within 0.99--1.03$\times$ that of the dense model
(\cref{tab:supp_reliab}).

\subsection{Cost--Utility Allocation}
\label{sec:allocator}

The trained model supports all configurations in
$\mathcal{C}$. We next decide which one to execute
under a given compute budget, via
validation-time utility calibration and per-sample
selection at inference (\cref{fig:overview}(c)--(d)).

\noindent\textbf{Validation-time calibration.}
We build a utility table over complete
modality-depth configurations (\cref{fig:overview}(c)). For each
$\bm{c}\in\mathcal{C}$, we compute its mean absolute error on $\mathcal{V}$:
\begin{equation}
  u(\bm{c})
  =\frac{1}{|\mathcal{V}|}
  \sum_{i\in\mathcal{V}}
  |\hat{y}_i(\bm{c})-y_i|,
  \label{eq:utility}
\end{equation}
where $u(\bm{c})$ is the utility of configuration
$\bm{c}$ (lower is better), $\mathcal{V}$ is the
validation set, and $|\mathcal{V}|$ is its size.
Unlike independent modality scores, $u(\bm{c})$
evaluates the fused prediction and therefore
captures interactions among modality depths.
The utility table is computed once from full-depth
encoder passes over $\mathcal{V}$ and reused for all
budgets and test samples.

\noindent\textbf{Per-sample computational cost.}
At inference, the first step in
\cref{fig:overview}(d) estimates the cost of each
exit from the input sequence lengths.
We model the FLOPs of one Transformer layer of
modality $m$ as
\begin{equation}
  \phi_m(n)=
  2\left(
    4nd_m^2+2n^2d_m+2nd_m d_{\mathrm{ff},m}
  \right),
  \label{eq:layercost}
\end{equation}
where $n$ is the number of input tokens, $d_m$ and
$d_{\mathrm{ff},m}$ are the hidden and feed-forward
widths of the encoder, the three terms count the
attention projections, the attention score and
value products, and the feed-forward network, and
the factor 2 converts multiply--accumulate
operations into FLOPs.
The cost of selecting exit $j$ for modality $m$
on sample $i$ is
\begin{equation}
  \mathrm{cost}_{i,m}^{\,j}
  =\ell_m^j\phi_m(n_{i,m})
  +e_{i,m}\mathbb{1}[j>0],
  \label{eq:cost}
\end{equation}
where $n_{i,m}$ is the input token count,
$\ell_m^j$ is the selected encoder depth,
$e_{i,m}$ is the input-side computational cost,
and $\mathbb{1}[\cdot]$ is the indicator function,
with $\ell_m^0=0$ for skipping.
We include $e_{i,m}$ in \method-skip and set it
to zero in the depth-only variant's budget
accounting, since input-side computation cannot
be avoided without skipping.
For modalities with multiple input streams,
such as utterance and context, we sum the costs
of their respective streams.

\begin{table*}[t]
  \centering
  \caption{\textbf{Main results on \dsmosi{} and \dsmosei{}.} \method{} retains competitive accuracy at lower measured FLOPs; skipping enables lower-cost operating points. One \method{} and one \method-skip model each serve all budgets. ADMN uses the same backbone with a budget-specific controller. \textbf{Bold}: lowest GFLOPs and latency at each budget. Mean over three seeds.}
  \label{tab:main}
  \fittable{\textwidth}{%
  \begin{tabular}{l ccccc cc c ccccc cc}
    \toprule
    & \multicolumn{7}{c}{\dsmosi} && \multicolumn{7}{c}{\dsmosei} \\
    \cmidrule{2-8}\cmidrule{10-16}
    Method & Acc-2\up & F1\up & Acc-7\up & MAE\dn & Corr\up & GFLOPs\dn & Lat.\,(ms)\dn && Acc-2\up & F1\up & Acc-7\up & MAE\dn & Corr\up & GFLOPs\dn & Lat.\,(ms)\dn \\
    \midrule
    Dense T+A+V & 88.77 & 88.79 & 47.81 & 0.595 & 0.868 & 139.2 & 56.8 && 87.91 & 87.95 & 56.48 & 0.488 & 0.818 & 172.9 & 56.5 \\
    \midrule
    ADMN, $\rho=0.8$ & 88.92 & 88.92 & 47.72 & 0.644 & 0.849 & 118.7 & 66.0 && 87.56 & 87.64 & 55.93 & 0.493 & 0.816 & 144.8 & 66.1 \\
    ADMN, $\rho=0.6$ & 89.38 & 89.36 & 49.75 & 0.616 & 0.854 & 98.2 & 63.5 && 87.90 & 87.95 & 56.39 & 0.486 & 0.815 & 122.2 & 60.5 \\
    ADMN, $\rho=0.4$ & 88.11 & 88.08 & 47.96 & 0.639 & 0.846 & 79.6 & 57.4 && 86.71 & 86.82 & 55.21 & 0.500 & 0.804 & 98.3 & 54.1 \\
    ADMN, $\rho=0.25$ & 84.50 & 84.54 & 40.62 & 0.773 & 0.776 & 65.8 & 48.4 && 86.02 & 86.07 & 52.79 & 0.536 & 0.770 & 80.8 & 44.9 \\
    \midrule
    \rowcolor{oursblue} \method{}, $\rho=0.8$ & 88.77 & 88.77 & 47.96 & 0.619 & 0.853 & 92.3 & 52.6 && 87.06 & 87.06 & 55.63 & 0.496 & 0.808 & 135.2 & 57.5 \\
    \rowcolor{oursblue} \method{}, $\rho=0.6$ & 88.67 & 88.67 & 47.57 & 0.622 & 0.852 & 86.9 & 49.0 && 87.07 & 87.05 & 55.29 & 0.499 & 0.806 & 109.7 & 50.5 \\
    \rowcolor{oursblue} \method{}, $\rho=0.4$ & 75.41 & 75.54 & 36.35 & 0.938 & 0.657 & 76.9 & 38.3 && 86.95 & 86.97 & 54.59 & 0.505 & 0.804 & 94.9 & 40.1 \\
    \rowcolor{oursblue} \method{}, $\rho=0.25$ & 49.59 & 47.37 & 17.54 & 1.558 & $-$0.018 & 66.2 & \textbf{23.7} && 84.47 & 84.44 & 50.67 & 0.566 & 0.749 & 81.3 & \textbf{24.0} \\
    \midrule
    \rowcolor{ourspurple} \method-skip, $\rho=0.8$ & 87.96 & 87.98 & 48.44 & 0.622 & 0.861 & \textbf{44.5} & \textbf{40.9} && 86.64 & 86.70 & 55.54 & 0.497 & 0.807 & \textbf{127.9} & \textbf{48.5} \\
    \rowcolor{ourspurple} \method-skip, $\rho=0.6$ & 88.11 & 88.13 & 48.15 & 0.614 & 0.864 & \textbf{28.5} & \textbf{37.6} && 86.80 & 86.89 & 55.74 & 0.499 & 0.807 & \textbf{99.1} & \textbf{43.5} \\
    \rowcolor{ourspurple} \method-skip, $\rho=0.4$ & 88.06 & 88.08 & 48.88 & 0.610 & 0.865 & \textbf{24.7} & \textbf{37.1} && 87.23 & 87.28 & 55.05 & 0.504 & 0.806 & \textbf{30.3} & \textbf{34.6} \\
    \rowcolor{ourspurple} \method-skip, $\rho=0.25$ & 86.43 & 86.48 & 47.47 & 0.637 & 0.850 & \textbf{23.7} & 35.9 && 87.24 & 87.29 & 55.13 & 0.505 & 0.806 & \textbf{29.3} & 34.0 \\
    \bottomrule
  \end{tabular}}
\end{table*}

\noindent\textbf{Budgeted configuration selection.}
Given the calibrated utility table and per-sample
costs, the second step in \cref{fig:overview}(d)
selects a configuration by solving
\begin{equation}
  \bm{c}_i^\star
  =\arg\min_{\bm{c}\in\mathcal{C}}u(\bm{c})
  \quad\mathrm{s.t.}\quad
  \sum_{m=1}^{M}\mathrm{cost}_{i,m}^{\,c_m}
  \le B_i,
  \label{eq:knapsack}
\end{equation}
where $\bm{c}_i^\star$ is the selected configuration,
$B_i=\rho C_i$ is the sample-specific budget,
and $C_i=\sum_m\mathrm{cost}_{i,m}^{\,J}$
is the modelled full-depth cost.
Ties are resolved in favor of the cheaper
configuration. With at most 125 combinations,
we solve this constrained selection exactly
by enumeration, taking approximately
$0.2$\,ms per sample on a CPU.

Finally, as shown in the third step of
\cref{fig:overview}(d), each encoder executes
only to its selected exit, and skipped
modalities are not encoded. Thus, the selected
configuration satisfies the modelled FLOPs
constraint by construction, and computation
is saved through actual early stopping rather
than by merely masking the unused layers.

The allocation can change across samples because
their input lengths affect configuration costs.
It is therefore \emph{cost-adaptive}, rather
than content-adaptive, since the utility table
is calibrated at the dataset level.
The allocator may also leave part of the budget
unused when additional depth does not reduce
validation error. Hence, one trained model and
utility table serve all budgets without retraining.

\section{Experiments}
\label{sec:experiments}

\subsection{Setup}
\label{sec:setup}

\noindent\textbf{Datasets and metrics.}
We evaluate on \dsmosi{}~\cite{mosi} (2,199 English video clips), \dsmosei{}~\cite{mosei} (22,856 clips), and \dssims{}~\cite{chsims} (2,281 Mandarin clips with unimodal sentiment labels), using the standard splits and MMSA evaluation protocol~\cite{mmsa}. We report binary accuracy and F1 (Acc-2 and F1), seven- or three-class accuracy (Acc-7 or Acc-3), mean absolute error (MAE), and the Pearson correlation (Corr) with the labels.

\noindent\textbf{Backbones and baselines.}
Our base model follows MMML~\cite{mmml}, using RoBERTa-large~\cite{roberta} and data2vec-audio~\cite{data2vec} for English, and Chinese RoBERTa-wwm~\cite{chinese-roberta-wwm} and HuBERT~\cite{hubert} for Mandarin. The standard visual stream uses a four-layer Transformer over pre-extracted Facet or OpenFace features. We compare against full-depth dense models and ADMN~\cite{admn}, reimplemented on the same backbones. For a vision-heavy setting, we replace the visual stream with a CLIP ViT-B/16~\cite{radford2021learning} operating on eight raw frames. All models use AdamW with batch size 8. Main results, ADMN, and CLIP experiments average three seeds.

\noindent\textbf{Compute measurement.}
The budget ratio $\rho$ constrains the modelled per-sample FLOPs in \cref{eq:cost}, whereas all reported GFLOPs measure end-to-end inference, including input-side processing, readouts, and fusion. Latency is measured on an NVIDIA A100 at batch size one with unpadded inputs. The depth-only and skip-enabled variants account for input-side computation differently in their constraints, so we compare their \emph{measured GFLOPs}, not their $\rho$ values.

\subsection{Main Results}
\label{sec:main}

\begin{figure*}[t]
  \centering
  \includegraphics[width=0.8\textwidth]{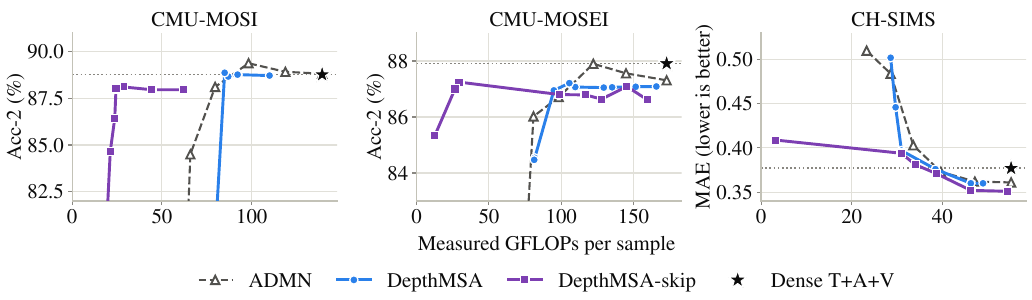}
  \caption{\textbf{Accuracy--efficiency trade-offs} on three MSA datasets. Each \method{} or \method-skip curve is obtained from one trained model evaluated at multiple budgets, whereas ADMN uses a controller for each budget. The skip-enabled model extends the feasible operating range to lower measured FLOPs. Lines connect operating points. Mean over three seeds.}
  \label{fig:pareto}
\end{figure*}

\noindent\textbf{Accuracy--efficiency trade-offs.}
\Cref{tab:main,fig:pareto} show that \method{} preserves competitive sentiment accuracy over a range of budgets while reducing inference cost. On \dsmosei{}, the depth-only model at $\rho=0.4$ retains 86.95\% Acc-2, versus 87.91\% for the dense model, with 45\% fewer measured FLOPs (94.9 against 172.9 GFLOPs) and 29\% lower latency (40.1 against 56.5\,ms). On \dsmosi{}, at $\rho=0.6$ it retains nearly the same Acc-2 (88.67 against 88.77) with 38\% fewer FLOPs (86.9 against 139.2 GFLOPs). The curves stay relatively flat across several operating points, indicating that full-depth computation is not always necessary to preserve sentiment prediction. This flexibility comes from one trained model per variant and dataset, not a separate model for every budget.

\noindent\textbf{The role of modality skipping.}
The depth-only model must execute at least one exit per modality, which imposes a minimum cost. In contrast, \method-skip can bypass an encoder and its input-side computation. At $\rho=0.4$, the skip-enabled variant uses 82\% fewer measured FLOPs than the dense model (24.7 against 139.2 GFLOPs on \dsmosi{}, 30.3 against 172.9 on \dsmosei{}) and stays within one percentage point of its Acc-2 (88.06 against 88.77, 87.23 against 87.91; \cref{tab:main}). It also avoids the cliff of the depth-only model on \dsmosi{} discussed next: skipping audio frees enough budget to keep text at full depth. Thus, skipping is not merely an additional exit: it extends the accuracy--efficiency trade-off into a regime that depth truncation alone cannot reach on these benchmarks.

\begin{figure*}[t]
  \centering
  \includegraphics[width=0.8\textwidth]{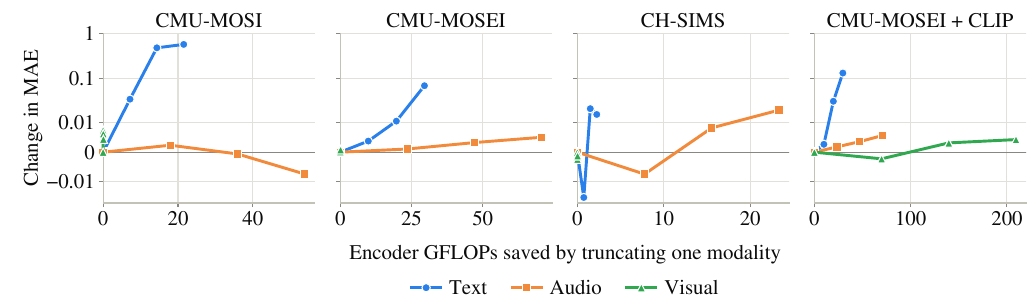}
  \caption{\textbf{Compute--utility landscape.} Each curve truncates one modality through exits 3, 2, and 1 while keeping the others at full depth. The horizontal axis shows encoder GFLOPs saved; the vertical axis shows the change in validation MAE relative to full depth (symmetric log scale). Larger savings with smaller error changes indicate greater depth redundancy. The modality offering this trade-off changes from audio with pre-extracted visual features to vision with CLIP. Mean over three seeds.}
  \label{fig:landscape}
\end{figure*}

\noindent\textbf{Why does the depth-only model fail at tight budgets on \dsmosi{}?}
Without skipping, each modality runs at least its first exit. On \dsmosi{}, unlike \dsmosei{}, the first two text exits are far worse than the third (\cref{fig:landscape}), and $\rho\le0.4$ pushes text below its third exit for some or all samples, also with CLIP (\cref{tab:clip}). The coarse exit grid, not the selection, causes the drop: at $\rho=0.4$, even a test-label oracle stays above 0.79 MAE (\cref{tab:supp_alloc_mosi_noskip}), whereas with $J=12$ exits the depth-only model keeps 87.80 and 85.37 Acc-2 at $\rho=0.4$ and $0.25$ (\cref{tab:supp_exits_J}).

\begin{table}[t]
  \centering
  \caption{\textbf{Main results on \dssims{}.} \method{} achieves a lower mean MAE than the dense model at a moderate budget, and comparable MAE to ADMN with lower measured latency. Mean over three seeds. FLOPs, latency and bold as in \cref{tab:main}.}
  \label{tab:sims}
  \fittable{\columnwidth}{%
  \begin{tabular}{l ccccc cc}
    \toprule
    Method & Acc-2\up & F1\up & Acc-3\up & MAE\dn & Corr\up & GFLOPs\dn & Lat.\dn \\
    \midrule
    Dense T+A+V & 79.43 & 79.72 & 69.51 & 0.377 & 0.661 & 55.2 & 21.2 \\
    \midrule
    ADMN, $\rho=0.8$ & 79.87 & 80.29 & 69.73 & 0.362 & 0.711 & 47.2 & 27.8 \\
    ADMN, $\rho=0.6$ & 80.89 & 81.12 & 69.07 & 0.373 & 0.688 & 39.7 & 25.9 \\
    ADMN, $\rho=0.4$ & 78.55 & 78.84 & 66.38 & 0.403 & 0.643 & 33.6 & 24.5 \\
    ADMN, $\rho=0.25$ & 74.25 & 72.40 & 60.69 & 0.484 & 0.528 & 28.6 & 19.1 \\
    \midrule
    \rowcolor{oursblue} \method{}, $\rho=0.8$ & 79.07 & 79.61 & 70.17 & 0.360 & 0.727 & 46.3 & 20.4 \\
    \rowcolor{oursblue} \method{}, $\rho=0.6$ & 77.68 & 78.32 & 69.37 & 0.376 & 0.691 & 38.5 & 18.5 \\
    \rowcolor{oursblue} \method{}, $\rho=0.4$ & 77.53 & 78.06 & 68.05 & 0.397 & 0.633 & 31.0 & 20.3 \\
    \rowcolor{oursblue} \method{}, $\rho=0.25$ & 68.49 & 68.18 & 56.82 & 0.502 & 0.401 & 28.6 & \textbf{10.5} \\
    \midrule
    \rowcolor{ourspurple} \method-skip, $\rho=0.8$ & 79.21 & 79.68 & 69.81 & 0.371 & 0.689 & \textbf{38.7} & \textbf{18.6} \\
    \rowcolor{ourspurple} \method-skip, $\rho=0.6$ & 78.04 & 78.51 & 68.27 & 0.394 & 0.636 & \textbf{30.9} & \textbf{17.7} \\
    \rowcolor{ourspurple} \method-skip, $\rho=0.4$ & 76.95 & 77.31 & 65.94 & 0.409 & 0.600 & \textbf{3.1} & \textbf{14.1} \\
    \rowcolor{ourspurple} \method-skip, $\rho=0.25$ & 76.95 & 77.31 & 65.94 & 0.409 & 0.600 & \textbf{3.1} & 14.1 \\
    \bottomrule
  \end{tabular}}
\end{table}

\noindent\textbf{Comparison with adaptive allocation.}
ADMN~\cite{admn} is our closest baseline because it also distributes encoder layers across modalities. The main difference appears at low measured compute (\cref{fig:pareto}): on \dsmosei{} at $\rho=0.4$, \method-skip achieves higher Acc-2 than ADMN (87.23 against 86.71) while using 3.2$\times$ fewer GFLOPs (30.3 against 98.3) and 36\% lower latency (34.6 against 54.1\,ms). On \dsmosi{}, it attains a lower MAE than ADMN at every budget with fewer FLOPs, \eg, 0.610 against 0.639 MAE at 24.7 against 79.6 GFLOPs for $\rho=0.4$. At moderate budgets, the two approaches are often comparable in accuracy, but \method{} can have lower inference latency without a budget-specific controller. These results suggest the benefit of our proposed joint depth supervision and cost-aware selection.

\noindent\textbf{Results on \dssims{}.}
\Cref{tab:sims} provides a contrasting case in which retaining all modalities is valuable, but full-depth encoding is still unnecessary: a model retrained without audio has a higher MAE than the dense model (0.409 against 0.377; \cref{tab:supp_main_sims}), so audio cannot simply be removed. At $\rho=0.8$, \method{} improves the mean MAE from 0.377 to 0.360 while reducing FLOPs by 16\% (46.3 against 55.2 GFLOPs). Its MAE is comparable to ADMN at a similar measured compute level (0.360 against 0.362 at 46.3 and 47.2 GFLOPs), with lower latency (20.4 against 27.8\,ms). The gain is concentrated in MAE and correlation rather than every classification metric, and the seed variation cautions against interpreting it as a universal improvement. More importantly, a shallower configuration can be preferable even when the budget permits deeper encoding. At $\rho\le0.4$, \method-skip keeps text at full depth and skips audio (3.1 GFLOPs), since even the first audio exit exceeds these budgets; hence the identical rows. Their MAE equals that of the model retrained without audio (0.409): the skip path matches a dedicated model, and audio matters here.

\subsection{Ablation Study}
\label{sec:ablation}

\noindent\textbf{Is combinatorial supervision necessary?}
\Cref{tab:ablation} isolates how the fusion head is trained while keeping the exit architecture fixed. When fusion is supervised only at full depth, performance deteriorates sharply as the budget tightens: on \dsmosei{} at $\rho=0.4$, MAE rises from 0.503 to 0.758. Having intermediate encoder features is therefore not enough; the fusion head must learn to combine features drawn from \emph{different} depths. Sampling one configuration per step performs substantially better than full-depth-only training, but full combinatorial supervision is more reliable at the tightest budgets, at negligible training overhead.

\begin{table}[t]
  \centering
  \caption{\textbf{Ablation of combinatorial anytime training} (MAE $\downarrow$). All methods use the same exit architecture. Full depth only supervises fusion at the deepest exits; one random samples one depth combination per sample and step. All combinations indicates ours, while the weight $w=0$ removes the full-depth supervision.}
  \label{tab:ablation}
  \fittable{\columnwidth}{%
  \begin{tabular}{l cccc c cccc}
    \toprule
    & \multicolumn{4}{c}{\dsmosei} && \multicolumn{4}{c}{\dssims} \\
    \cmidrule{2-5}\cmidrule{7-10}
    Fusion supervision & full & $\rho{=}0.6$ & $\rho{=}0.4$ & $\rho{=}0.25$ && full & $\rho{=}0.6$ & $\rho{=}0.4$ & $\rho{=}0.25$ \\
    \midrule
    \rowcolor{oursblue} All combinations & \textbf{0.494} & \textbf{0.500} & \textbf{0.503} & \textbf{0.554} && 0.378 & \textbf{0.365} & 0.392 & 0.504 \\
    One random & \textbf{0.494} & 0.501 & 0.507 & 0.570 && \textbf{0.370} & 0.373 & \textbf{0.388} & 0.517 \\
    Full depth only & 0.501 & 0.518 & 0.758 & 0.806 && 0.374 & 0.395 & 0.403 & 0.592 \\
    \midrule
    All, $w{=}0$ & 0.499 & 0.502 & 0.505 & \textbf{0.554} && 0.385 & 0.367 & 0.405 & \textbf{0.433} \\
    \bottomrule
  \end{tabular}}
\end{table}

\noindent\textbf{Does full-depth emphasis help?}
The extra full-depth loss balances performance at the deepest configuration against robustness at shallow configurations. Removing it ($w=0$) slightly worsens several full- and moderate-budget settings, but can improve MAE at the tightest budgets, particularly on \dssims{} (\cref{tab:ablation}). Thus, no single $w$ is optimal: emphasizing full-depth prediction protects the default configuration, whereas anytime robustness favors less emphasis.

\noindent\textbf{Does the allocation rule matter?}
On the same trained models, uniform truncation can severely hurt prediction because it reduces sensitive modalities along with redundant ones; on \dsmosi{} at $\rho=0.6$, it yields 1.376 MAE, versus 0.622 for our cost--utility selection (\cref{tab:supp_alloc_mosi_noskip}). Greedy text-first or random priorities are more competitive, but cannot account for the joint utility of a configuration and may spend additional FLOPs without improving validation error. The benefit of our allocator is not a preference for text, but selecting an effective combination under each budget.

\subsection{Further Analysis and Discussion}
\label{sec:analysis}

\begin{table*}[t]
  \centering
  \caption{\textbf{Results with a CLIP visual encoder.} CLIP ViT-B/16 processes eight raw frames and accounts for 74\% and 68\% of encoder FLOPs on \dsmosi{} and \dsmosei{}, respectively. The same depth-only model serves all budgets within each dataset. $\Delta$ is the change in measured end-to-end GFLOPs relative to the corresponding full-depth configuration. Mean over three seeds.}
  \label{tab:clip}
  \fittable{\textwidth}{%
  \begin{tabular}{l cccc ccc c cccc ccc}
    \toprule
    & \multicolumn{7}{c}{\dsmosi} && \multicolumn{7}{c}{\dsmosei} \\
    \cmidrule{2-8}\cmidrule{10-16}
    Budget & Acc-2\up & F1\up & MAE\dn & Corr\up & GFLOPs\dn & $\Delta$ & Lat.\,(ms)\dn && Acc-2\up & F1\up & MAE\dn & Corr\up & GFLOPs\dn & $\Delta$ & Lat.\,(ms)\dn \\
    \midrule
    \method{}, full depth & 86.28 & 86.31 & 0.761 & 0.817 & 420.1 & -- & 82.2 && 87.41 & 87.47 & 0.490 & \textbf{0.814} & 453.8 & -- & 83.7 \\
    \rowcolor{oursblue} \method{}, $\rho=0.8$ & \textbf{88.97} & \textbf{88.97} & \textbf{0.661} & \textbf{0.841} & 187.3 & $-55\%$ & 60.1 && \textbf{87.64} & \textbf{87.70} & \textbf{0.489} & 0.813 & 356.4 & $-21\%$ & 71.1 \\
    \rowcolor{oursblue} \method{}, $\rho=0.6$ & \textbf{88.97} & \textbf{88.97} & \textbf{0.661} & \textbf{0.841} & 187.3 & $-55\%$ & 60.2 && 87.34 & 87.41 & 0.495 & 0.809 & 249.2 & $-45\%$ & 65.1 \\
    \rowcolor{oursblue} \method{}, $\rho=0.4$ & 88.52 & 88.51 & 0.670 & 0.839 & 167.1 & $-60\%$ & 58.6 && 87.40 & 87.44 & 0.497 & 0.808 & 197.6 & $-56\%$ & 58.3 \\
    \rowcolor{oursblue} \method{}, $\rho=0.25$ & 46.90 & 43.33 & 1.543 & $-$0.080 & 137.9 & $-67\%$ & 29.0 && 81.43 & 81.16 & 0.619 & 0.692 & 152.9 & $-66\%$ & 29.6 \\
    \bottomrule
  \end{tabular}}
\end{table*}

\begin{table*}[t]
  \centering
  \caption{\textbf{Different encoder combinations} (text / audio / visual). For each combination, one \method{} model is evaluated at full depth ($\rho=1.0$) and at the budget $\rho\in\{0.8,0.6,0.4,0.25\}$ with the highest Acc-2 (shaded). Parentheses: change in measured GFLOPs relative to full depth. Mean over three seeds.}
  \label{tab:backbones}
  \fittable{\textwidth}{%
  \begin{tabular}{l c ccccc cc c c ccccc cc}
    \toprule
    & \multicolumn{8}{c}{\dsmosi} && \multicolumn{8}{c}{\dsmosei} \\
    \cmidrule{2-9}\cmidrule{11-18}
    Encoders & $\rho$ & Acc-2\up & F1\up & Acc-7\up & MAE\dn & Corr\up & GFLOPs\dn & Lat.\,(ms)\dn && $\rho$ & Acc-2\up & F1\up & Acc-7\up & MAE\dn & Corr\up & GFLOPs\dn & Lat.\,(ms)\dn \\
    \midrule
    RoBERTa-L / data2vec / Facet & 1.0 & 88.62 & 88.62 & 48.15 & 0.606 & 0.860 & 139.2 & 61.6 && 1.0 & 87.14 & 87.13 & 55.71 & 0.495 & 0.808 & 172.9 & 62.5 \\
    \rowcolor{oursblue}  & $0.8$ & 88.77 & 88.77 & 47.96 & 0.619 & 0.853 & \textbf{92.3} {\scriptsize($-34\%$)} & 52.6 && $0.6$ & 87.07 & 87.05 & 55.29 & 0.499 & 0.806 & \textbf{109.7} {\scriptsize($-37\%$)} & 50.5 \\
    \midrule
    BERT-base / wav2vec\,2.0 / Facet & 1.0 & 85.37 & 85.39 & 43.30 & 0.735 & 0.796 & 120.8 & 42.6 && 1.0 & 86.26 & 86.29 & 54.91 & 0.517 & 0.786 & 148.3 & 43.5 \\
    \rowcolor{oursblue}  & $0.4$ & 85.42 & 85.43 & 42.52 & 0.737 & 0.792 & \textbf{68.0} {\scriptsize($-44\%$)} & 30.8 && $0.6$ & 86.23 & 86.27 & 54.31 & 0.521 & 0.783 & \textbf{105.4} {\scriptsize($-29\%$)} & 35.3 \\
    \midrule
    RoBERTa-L / data2vec / CLIP & 1.0 & 86.28 & 86.31 & 39.50 & 0.761 & 0.817 & 420.1 & 82.2 && 1.0 & 87.41 & 87.47 & 56.14 & 0.490 & 0.814 & 453.8 & 83.7 \\
    \rowcolor{oursblue}  & $0.8$ & 88.97 & 88.97 & 45.72 & 0.661 & 0.841 & \textbf{187.3} {\scriptsize($-55\%$)} & 60.1 && $0.8$ & 87.64 & 87.70 & 56.22 & 0.489 & 0.813 & \textbf{356.4} {\scriptsize($-21\%$)} & 71.1 \\
    \bottomrule
  \end{tabular}}
\end{table*}

\noindent\textbf{Where does additional depth matter?}\label{sec:observations}
\Cref{fig:landscape} examines the change in validation MAE when one modality is truncated while the others remain at full depth. With pre-extracted visual features, audio dominates encoder FLOPs but its deepest layers offer little additional predictive value. On \dsmosei{}, truncating audio to a quarter of its depth saves 71 GFLOPs for only 0.005 MAE, whereas the same truncation of text saves 30 GFLOPs at a 0.068 MAE cost (\cref{fig:teaser}a). The point is not that audio lacks useful sentiment cues; it is that \emph{modality importance and the marginal utility of deeper encoding are different quantities}. This explains why uniform depth reduction is ineffective for MSA.

\begin{figure}[t]
  \centering
  \includegraphics[width=0.9\columnwidth]{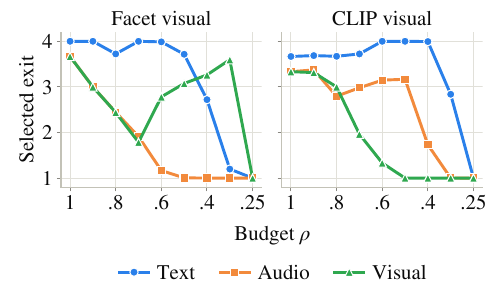}
  \caption{\textbf{Allocation reversal under different visual encoders.} Mean selected exit on \dsmosei{} at each budget (three seeds). With inexpensive Facet visual features, audio is truncated first; with CLIP, the allocator first reduces visual depth.}
  \label{fig:alloc}
\end{figure}

\noindent\textbf{What happens when vision becomes expensive?}\label{sec:clip}
To examine whether the allocation depends on the modality itself or on its cost--utility structure, we replace pre-extracted visual features with a CLIP ViT-B/16 on raw frames (\cref{tab:clip}). Vision then accounts for 74\% and 68\% of the encoder FLOPs on \dsmosi{} and \dsmosei{}. On \dsmosei{}, \method{} at $\rho=0.4$ reduces measured FLOPs by 56\% while retaining the full-depth Acc-2 (87.40 against 87.41). On \dsmosi{}, shallower CLIP exits are also competitive at much lower compute, although the high variance at full depth precludes a claim of consistent improvement. Crucially, \cref{fig:alloc} shows an \emph{allocation reversal}: with Facet, audio is truncated first, whereas with CLIP, vision receives the shallower exits. Text stays protected in both settings. This supports a cost--utility interpretation rather than a fixed audio-first heuristic.

\noindent\textbf{How do encoder combinations affect DepthMSA?}
The encoders in \cref{tab:backbones} cover common MSA choices: BERT with Facet features, the standard setting~\cite{misa,selfmm,mmim,almt}; RoBERTa-large with data2vec~\cite{mmml}; the widely used wav2vec\,2.0~\cite{wav2vec2}; and CLIP~\cite{radford2021learning} for raw frames. Across all combinations, the best budget matches or exceeds the full-depth accuracy (at most 0.07 Acc-2 lower) with 21--55\% fewer measured FLOPs. The depth redundancy is therefore not specific to one large model: with the smaller BERT-base and wav2vec\,2.0 encoders, \method{} at $\rho=0.4$ keeps the full-depth Acc-2 on \dsmosi{} (85.42 against 85.37) with 44\% fewer FLOPs. A truncated strong encoder can also replace a smaller one run to full depth: on \dsmosei{}, RoBERTa-large at $\rho=0.6$ is more accurate than the base encoders at full depth (87.07 against 86.26 Acc-2) and cheaper (109.7 against 148.3 GFLOPs).

\noindent\textbf{What determines sample-specific decisions?}
As in \cref{sec:allocator}, the utility table is shared across test samples, and only their input lengths change the configuration costs. Thus, the allocator is \emph{cost-adaptive}, not content-adaptive. Across settings, validation and test rankings of configurations are strongly correlated (Spearman 0.85--0.99), and a utility table calibrated on 57 validation examples (25\% of \dsmosi{}) is within 0.006 MAE of the full one (\cref{app:reliab}). Estimating per-sample modality importance or stopping by confidence does not improve the trade-off on these benchmarks (\cref{tab:supp_stop,tab:supp_alloc_mosi_noskip}). These results suggest that, for the clean MSA settings studied, exploiting the stable cost--utility landscape can be more reliable than adding an input-dependent controller.

\section{Conclusion}
\label{sec:conclusion}

We present \method{}, an efficient MSA framework that enables flexible computation across modality encoders under varying budgets. By combining combinatorial anytime training with cost-aware depth allocation, \method{} supports heterogeneous encoding depths and selects configurations according to their predictive utility and computational cost. Experiments across three benchmarks and different encoder architectures demonstrate substantial FLOPs reductions while maintaining competitive sentiment prediction performance. Further analysis reveals that computational cost and depth-dependent predictive utility are often misaligned, and the preferred allocation changes with the encoder architecture. These findings highlight the importance of allocating multimodal computation according to its predictive benefit rather than uniformly processing every modality to full depth.

\section*{LLM Usage Statement}
We used large language models to assist with manuscript editing, implementation of experimental and analysis scripts, and preparation of figures. All AI-assisted materials were examined and validated by the authors, who take full responsibility for the scientific content, experimental findings, and final presentation of this work.

{\small
\bibliographystyle{ieeenat_fullname}
\bibliography{main}
}

\clearpage
\maketitlesupplementary
\appendix
\setcounter{table}{0}
\renewcommand{\thetable}{S\arabic{table}}
\renewcommand{\theHtable}{appendix.\arabic{table}}
\setcounter{figure}{0}
\renewcommand{\thefigure}{S\arabic{figure}}
\renewcommand{\theHfigure}{appendix.\arabic{figure}}

This supplement gives the cost model (\cref{app:flops}), implementation details (\cref{app:impl}), the checks behind the evaluation protocol (\cref{app:checks}), the reliability of the training-free allocator and the cost of combinatorial anytime training (\cref{app:reliab}), the complete results at every budget (\cref{app:results}), and a qualitative analysis (\cref{app:qual}).

\section{Cost Model}
\label{app:flops}

\noindent\textbf{Layers.} We count FLOPs as twice the multiply-accumulate operations. A Transformer layer of width $d$ and feed-forward width $d_{\mathrm{ff}}$ on $n$ tokens costs $\phi(n)=2(4nd^{2}+2n^{2}d+2nd\,d_{\mathrm{ff}})$. The cost of exit $j$ of a modality is the number of layers up to that exit times the sum of $\phi$ over its streams (utterance and context), evaluated at the valid length of each stream, so padding is not counted. For CLIP, the per-layer cost is the number of frames times the cost of one layer on 197 tokens.

\noindent\textbf{Input-side compute.} In \method-skip, the cost of a modality includes the compute before its first layer: for audio, the convolutional feature extractor (a layer with output length $L$, kernel $k$ and channels $c_{\text{in}},c_{\text{out}}$ costs $2Lc_{\text{out}}c_{\text{in}}k$), the feature projection and the positional convolution; for CLIP, the patch embedding of each frame. On \dsmosei{} this adds about 56 GFLOPs per sample to the audio stream, which can only be saved by skipping it.

\noindent\textbf{Per-sample cost.} \Cref{tab:supp_cost} lists the mean full-depth cost per modality. Relative to one text layer, one audio layer costs about five times as much on \dsmosi{} and \dsmosei{} and about ten times as much on \dssims{}.

\begin{table}[h]
  \centering
  \caption{\textbf{Mean encoder cost at full depth} (GFLOPs per test sample, without input-side compute) and its split over text / audio / visual.}
  \label{tab:supp_cost}
  \fittable{\columnwidth}{%
  \begin{tabular}{l ccc c}
    \toprule
    Dataset, encoders & Text & Audio & Visual & Share T / A / V (\%) \\
    \midrule
    \dsmosi{} (RoBERTa-L, data2vec, Facet) & 28.7 & 71.8 & 0.09 & 28.6 / 71.3 / 0.1 \\
    \dsmosei{} (same) & 39.4 & 94.3 & 0.18 & 28.4 / 71.5 / 0.1 \\
    \dssims{} (RoBERTa-wwm, HuBERT, OpenFace) & 3.0 & 31.1 & 0.04 & 9.3 / 90.6 / 0.1 \\
    \dsmosi{}, BERT-base + wav2vec\,2.0 & 8.1 & 71.8 & 0.09 & 10.4 / 89.5 / 0.1 \\
    \dsmosi{}, CLIP visual & 28.7 & 71.8 & 279.2 & 7.4 / 18.5 / 74.1 \\
    \dsmosei{}, CLIP visual & 39.4 & 94.3 & 279.2 & 9.3 / 22.7 / 68.0 \\
    \bottomrule
  \end{tabular}}
\end{table}

\noindent\textbf{Feasibility floor.} Without the skip level, every modality runs its first exit, which is a quarter of its depth, so the cheapest combination costs exactly $C_i/4$ for every sample and budgets below 0.25 are infeasible. With the skip level the cheapest combination costs nothing, and every budget is feasible.

\section{Implementation Details}
\label{app:impl}

\noindent\textbf{Base model.} Text and audio are each encoded twice with the same encoder, once for the current utterance and once for its context (the two preceding utterances for text and one for audio on \dsmosi{} and \dsmosei{}; no context on \dssims{}), and the two features are concatenated. The text feature is a pooler (linear layer and $\tanh$) on the first token; the audio and visual features are masked means followed by layer normalisation. The fusion head is an MLP with hidden sizes 2048 and 1024, ReLU and dropout 0.3. \Cref{tab:supp_impl} lists the remaining settings.

\begin{table}[h]
  \centering
  \caption{\textbf{Implementation details.}}
  \label{tab:supp_impl}
  \fittable{\columnwidth}{%
  \begin{tabular}{ll}
    \toprule
    Exits & text: layers 6/12/18/24 (RoBERTa-L) or 3/6/9/12; \\
          & audio: 3/6/9/12; visual: 1/2/3/4 (CLIP: 3/6/9/12) \\
    Optimiser & AdamW, lr $5{\times}10^{-6}$ (\dssims{}: $10^{-5}$) \\
    Batch size, epochs & 8, 20 (early stopping, patience 8) \\
    Model selection & lowest validation $\mathcal{L}_{\text{all}}$ \\
    Loss weight & $w=1$; all losses are L1 \\
    Text input & at most 96 tokens (Mandarin: 64) \\
    Audio input & 16\,kHz waveform, at most 6\,s; frozen CNN extractor \\
    Visual input & Facet (20-d / 35-d), OpenFace (709-d), \\
                 & or 8 uniformly sampled $224{\times}224$ frames for CLIP \\
    Visual pool & 4 pre-norm layers, $d=128$, 4 heads, $d_{\mathrm{ff}}=512$ \\
    Unimodal labels & used for $\mathcal{L}_{\text{uni}}$ on \dssims{} \\
    Seeds & 1, 2, 3 (main results, models without audio or text only, ADMN, CLIP, BERT-base); \\
      & 1 (ablations); \dsmosi{} dense T: 2, 3, 4 \\
    Efficiency & batch size 1, unpadded, first 200 test samples, \\
               & 10 warm-up runs, NVIDIA A100 (40\,GB) \\
    \bottomrule
  \end{tabular}}
\end{table}

\noindent\textbf{Allocation baselines.} All rules are applied to the same trained model. \emph{Text-first} raises the depth of the modalities greedily in the fixed order text, audio, visual while the cap allows. \emph{Learned importance} does the same in the order given by a head that predicts a per-sample importance for each modality from the features of the first exit; in all our runs it ranks text first and therefore coincides with text-first. \emph{Uniform} gives all modalities the deepest common exit that fits. \emph{Random} uses a random order per sample. The \emph{oracle} selects, for each test sample, the feasible combination with the lowest error on that sample; it uses the test labels and is optimistic by construction, since it selects among up to 125 noisy predictions.

\noindent\textbf{ADMN baseline.} We port ADMN~\cite{admn} to our backbones. Stage 1 fine-tunes the backbone with LayerDrop using exactly our training configuration (learning rate, at most 20 epochs, early stopping and seed), so that both methods have the same training budget for the backbone. Stage 2 trains one controller per budget with ADMN's official settings (Adam, learning rate $10^{-3}$, 10 epochs), about two GPU-hours per budget on \dsmosei{}. The budget is the same per-sample cap on the modelled encoder cost as ours, and ADMN never exceeds it. On \dsmosi{}, stage 1 had not fully converged within 20 epochs on two of three seeds (validation MAE 0.665 and 0.720 against about 0.62 for the dense model); we did not extend it, to keep the training budget equal. On \dsmosei{} it converged on all seeds.

\noindent\textbf{Comparison with published models.} \Cref{tab:supp_sota} places our models among published MSA models. Our dense model is on par with MMML with context on \dsmosei{} and about one point below it on \dsmosi{}. On \dssims{}, our reimplementation with the encoders of \cref{sec:setup} is below the numbers reported for MMML (Acc-2 82.93, MAE 0.332). All comparisons in the paper are against our own dense model, trained under the same protocol as \method{}.

\begin{table}[h]
  \centering
  \caption{\textbf{Comparison with published MSA models} (numbers from the respective papers; Acc-2 is negative/positive). FLOPs are relative to our dense model (\dsmosi{} / \dsmosei{}). Our aim is not a higher accuracy but the same accuracy at a lower and adjustable cost.}
  \label{tab:supp_sota}
  \fittable{\columnwidth}{%
  \begin{tabular}{l ccc c ccc c}
    \toprule
    & \multicolumn{3}{c}{\dsmosi} && \multicolumn{3}{c}{\dsmosei} & \\
    \cmidrule{2-4}\cmidrule{6-8}
    Method & Acc-2 & MAE & Corr && Acc-2 & MAE & Corr & FLOPs \\
    \midrule
    MISA~\cite{misa} & 83.4 & 0.783 & 0.761 && 85.5 & 0.555 & 0.756 & -- \\
    Self-MM~\cite{selfmm} & 85.98 & 0.713 & 0.798 && 85.17 & 0.530 & 0.765 & -- \\
    MMIM~\cite{mmim} & 86.06 & 0.700 & 0.800 && 85.97 & 0.526 & 0.772 & -- \\
    UniMSE~\cite{unimse} & 86.9 & 0.691 & 0.809 && 87.50 & 0.523 & 0.773 & -- \\
    ALMT~\cite{almt} & 86.43 & 0.683 & 0.805 && 86.79 & 0.526 & 0.779 & -- \\
    KuDA~\cite{kuda} & 86.43 & 0.705 & 0.795 && 86.46 & 0.529 & 0.776 & -- \\
    MMML~\cite{mmml} & 88.16 & 0.643 & 0.838 && 86.73 & 0.517 & 0.791 & -- \\
    MMML + context~\cite{mmml} & 89.69 & 0.583 & 0.869 && 88.02 & 0.492 & 0.814 & -- \\
    \midrule
    Dense T+A+V (ours) & 88.77 & 0.595 & 0.868 && 87.91 & 0.488 & 0.818 & 100\% \\
    \rowcolor{oursblue} \method{}, $\rho=0.6$ & 88.67 & 0.622 & 0.852 && 87.07 & 0.499 & 0.806 & 62 / 63\% \\
    \rowcolor{ourspurple} \method-skip, $\rho=0.4$ & 88.06 & 0.610 & 0.865 && 87.23 & 0.504 & 0.806 & 18 / 18\% \\
    \bottomrule
  \end{tabular}}
\end{table}

\section{Evaluation Protocol and Checks}
\label{app:checks}

One full-depth pass over a split yields the features of all exits and hence the predictions of all combinations, which we cache. Every allocation rule at every budget is then evaluated exactly from this cache; the utility table is computed on the validation cache and all reported numbers on the test cache. We verify the following: (i) each pool run to its last layer reproduces the forward pass of the original pretrained encoder exactly; (ii) the analytic layer costs equal the output of a FLOP counter; (iii) every allocation rule stays within the cap for every sample; (iv) the deployment path, which really stops each encoder at its exit, reproduces the cached predictions up to $5{\times}10^{-6}$.

\section{Allocator Reliability and Training Cost}
\label{app:reliab}

\noindent\textbf{The cap holds.} For every run (all seeds, three datasets, with and without skipping, CLIP), 100.0\% of the test samples stay within the cap at every budget; the largest ratio of used to allowed modelled cost is 1.000.

\noindent\textbf{The utility table transfers from validation to test.} Ranking the 64 (125 with skipping) combinations by validation MAE and by test MAE gives a Spearman correlation of 0.85--0.99 over the ten run groups: on \dsmosi{} 0.93, 0.85 with skipping, 0.98 with CLIP and 0.97 with the base encoders; on \dsmosei{} 0.96, 0.98, 0.99 and 0.96; on \dssims{} 0.97 and 0.97. 75--92\% of the combinations are dominated in cost and validation error, so the allocator effectively chooses among 5--19 candidates.

\noindent\textbf{A small calibration set suffices.} We re-estimate the utility table from random subsets of the validation split (40 draws per size) and re-run the allocator on the full test split. With 25\% of the validation split (57 samples on \dsmosi{}) the test MAE at $\rho=0.6$ is within 0.006 of that obtained with the full split, and the regret against a table estimated on the test split itself is at most 0.005 on \dsmosei{} and \dssims{} and at most 0.014 on \dsmosi{} over all budgets. Only very small sets (11 samples) are unreliable.

\noindent\textbf{Cost of the allocator.} The allocator has no trainable parameters. Its only cost is one full-depth pass over the validation split to fill the utility table, done once for all budgets; at inference it selects a configuration by enumeration in about 0.2\,ms per sample on a CPU.

\noindent\textbf{Training cost of the model.} Combinatorial anytime training does not slow training down: training time per step and peak training memory are within 0.99--1.03$\times$ of those of dense training for every variant on all three datasets (\cref{tab:supp_reliab}, bottom), because the encoders run once per step and only the small fusion head is evaluated on every combination.

\begin{table}[h]
  \centering
  \caption{\textbf{Allocator reliability and training cost.} Top: test MAE at $\rho=0.6$ when the utility table is estimated from a fraction of the validation split (mean $\pm$ std over 40 draws and three seeds; 10\% and 25\% are 23 and 57 samples on \dsmosi{}), against the full split and a table estimated on the test split (oracle table); Spearman correlation between validation and test MAE over all combinations. Bottom: cost of training the model (not the allocator), as time per step and peak training memory relative to dense training (A100, identical batches; memory: largest ratio over the three datasets).}
  \label{tab:supp_reliab}
  \fittable{\columnwidth}{%
  \begin{tabular}{l ccccc}
    \toprule
    Run & 10\% & 25\% & 100\% & oracle table & Spearman \\
    \midrule
    \dsmosi{} & 0.640{\scriptsize$\pm$0.045} & 0.628{\scriptsize$\pm$0.015} & 0.622 & 0.620 & 0.932 \\
    \dsmosei{} & 0.501{\scriptsize$\pm$0.006} & 0.499{\scriptsize$\pm$0.003} & 0.499 & 0.497 & 0.955 \\
    \dssims{} & 0.391{\scriptsize$\pm$0.025} & 0.382{\scriptsize$\pm$0.013} & 0.376 & 0.376 & 0.970 \\
    \midrule
    & & \multicolumn{3}{c}{time per step} & \\
    \cmidrule{3-5}
    Training variant & fusion evals & \dsmosi{} & \dsmosei{} & \dssims{} & memory \\
    \midrule
    Dense T+A+V & 1 & 1.00$\times$ & 1.00$\times$ & 1.00$\times$ & 1.00$\times$ \\
    Full depth only & 64 & 0.99$\times$ & 1.01$\times$ & 1.01$\times$ & 1.01$\times$ \\
    One random combination & 64 & 0.99$\times$ & 1.01$\times$ & 1.01$\times$ & 1.01$\times$ \\
    \method{} (all 64) & 64 & 0.99$\times$ & 1.01$\times$ & 1.01$\times$ & 1.02$\times$ \\
    \method-skip (all 125) & 125 & 1.00$\times$ & 1.02$\times$ & 1.03$\times$ & 1.03$\times$ \\
    \bottomrule
  \end{tabular}}
\end{table}

\section{Complete Results}
\label{app:results}

\noindent\textbf{All budgets.} \Cref{tab:supp_main_mosi,tab:supp_main_mosei,tab:supp_main_sims} report every metric at every budget, with standard deviations. Consecutive budgets often select the same combination and then give identical rows. In \method-skip the standard deviation of the GFLOPs is large at high budgets on \dsmosi{}, because the seeds differ in whether audio is already skipped there.

\noindent\textbf{Allocation rules.} \Cref{tab:supp_alloc_mosi_noskip,tab:supp_alloc_mosi_skip,tab:supp_alloc_mosei_noskip,tab:supp_alloc_mosei_skip,tab:supp_alloc_sims_noskip,tab:supp_alloc_sims_skip} compare the allocation rules at every budget, with and without skipping, and include the oracle.

\begin{figure}[!t]
  \centering
  \includegraphics[width=0.93\columnwidth]{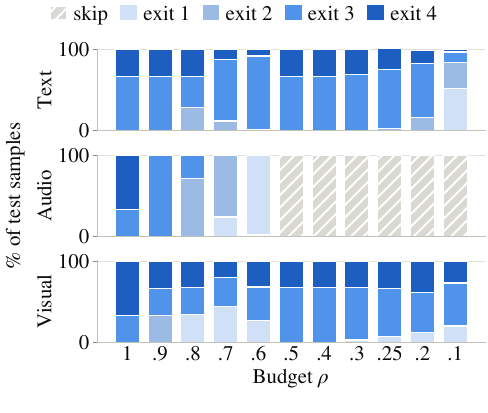}\\[2pt]
  \includegraphics[width=0.93\columnwidth]{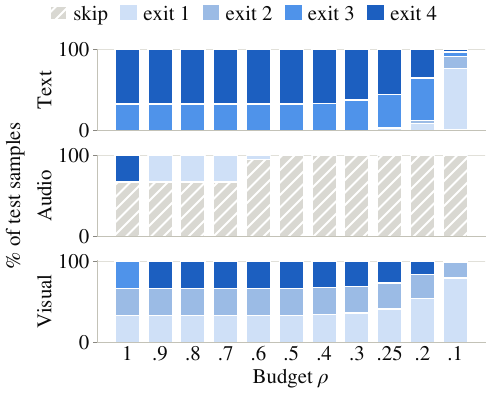}\\[2pt]
  \includegraphics[width=0.93\columnwidth]{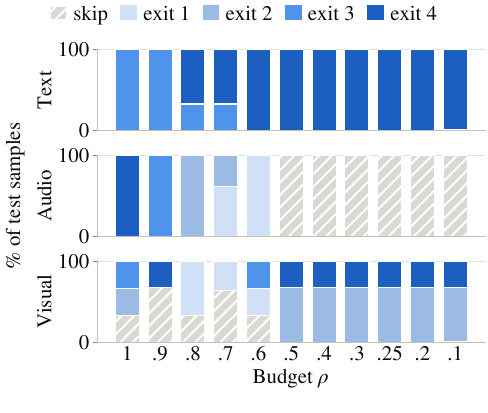}
  \caption{\textbf{Levels selected by \method-skip} on \dsmosei{}, \dsmosi{} and \dssims{} (top to bottom): share of test samples at each level, per modality, as the budget decreases (mean of 3 seeds).}
  \label{fig:supp_exits}
\end{figure}

\noindent\textbf{Training ablation.} \Cref{tab:supp_abl_mosi,tab:supp_abl_mosei,tab:supp_abl_sims} give the training ablation on all three datasets. Supervising the fusion head on mixed depths is necessary at low budgets on every dataset. Supervising all combinations is better than sampling one combination per step at every budget on \dsmosi{} and at $\rho\le0.3$ on \dsmosei{} and \dssims{}, while the two are close at higher budgets. The full-depth emphasis is a trade-off: on \dsmosi{}, and on \dssims{} at $\rho\le0.3$, the variant without it is more accurate at low budgets.

\noindent\textbf{Other encoders.} \Cref{tab:supp_bb_mosi,tab:supp_bb_mosei} give all metrics for the CLIP study and for smaller text and speech encoders (BERT-base~\cite{devlin2019bert} and wav2vec\,2.0~\cite{wav2vec2}; depth-only variant, 3 seeds; on \dsmosi{} also ADMN). With the smaller encoders the accuracy is again flat down to \mbox{$\rho=0.4$}. Truncating a strong model can be preferable to deploying a small one: on \dsmosei{}, RoBERTa-large at $\rho=0.4$ is more accurate than the base encoders at full depth (86.95 against 86.26 Acc-2, mean of 3 seeds) and cheaper (94.9 against 148.3 GFLOPs).

\noindent\textbf{Budget-free stopping.} \Cref{tab:supp_stop} compares rules that need no budget. Each modality stops when a criterion is met: \emph{fused} when the last exit changed the fused prediction by less than a threshold, \emph{unimodal} when the modality's own prediction has converged, and \emph{sample} when all modalities stop together, as in early exit for a single network. \emph{Fixed} uses one combination for all samples. Thresholds and the fixed combination are chosen on validation for a target compute. Stopping per modality is better than stopping per sample on \dsmosi{} and \dssims{}, and a fixed combination is as good as or better than every stopping rule.

\begin{table}[h]
  \centering
  \caption{\textbf{Budget-free stopping}. Test MAE (fraction of the full encoder FLOPs used) of the configuration selected on validation for a target compute.}
  \label{tab:supp_stop}
  \fittable{\columnwidth}{%
  \begin{tabular}{l c ccccc}
    \toprule
    Dataset & target & fixed & fused & unimodal & sample & random \\
    \midrule
    \dsmosei & 0.6 & 0.500 (0.46) & 0.498 (0.59) & 0.498 (0.60) & 0.501 (0.56) & 0.505 (0.50) \\
             & 0.8 & 0.494 (0.75) & 0.494 (0.77) & 0.495 (0.80) & 0.494 (0.75) & 0.494 (0.78) \\
    \dsmosi  & 0.6 & 0.626 (0.47) & 1.053 (0.56) & 1.006 (0.58) & 1.323 (0.57) & 1.228 (0.60) \\
             & 0.8 & 0.626 (0.47) & 0.634 (0.64) & 0.634 (0.64) & 0.678 (0.75) & 0.671 (0.78) \\
    \dssims  & 0.6 & 0.365 (0.52) & 0.391 (0.55) & 0.377 (0.59) & 0.447 (0.54) & 0.450 (0.53) \\
             & 0.8 & 0.351 (0.75) & 0.351 (0.75) & 0.354 (0.76) & 0.356 (0.75) & 0.356 (0.78) \\
    \bottomrule
  \end{tabular}}
\end{table}

\noindent\textbf{Selected levels.} \Cref{fig:supp_exits} shows the levels selected by \method-skip. Audio is skipped entirely from $\rho=0.5$ on \dsmosi{} and \dsmosei{}, while text keeps its third or fourth exit for most samples until very tight budgets; on \dssims{}, audio is kept at a shallow exit down to $\rho=0.6$. Shares in thirds reflect seeds that select different combinations; within a seed, almost all samples receive the same combination except at tight budgets.

\noindent\textbf{Number of exits.} \Cref{tab:supp_exits_J} varies the number of exits per modality $J$ on \dsmosi{} and \dssims{} (the four-layer visual stream has at most four distinct exits). Without skipping, a finer grid of exits helps at tight budgets: on \dsmosi{} at $\rho=0.25$, $J=12$ keeps 85.37 Acc-2, where $J=4$ and $J=6$ drop to 49.59 and 53.20, because audio can be cut to one twelfth of its depth, which leaves text at a usable depth. A coarse grid is the weakest choice: $J=2$ has no feasible configuration below $\rho=0.5$ and a higher error on \dsmosi{} (0.852 MAE at full depth). With skipping, all values of $J$ behave alike at low budgets, where the allocator skips audio and keeps text deep (87.65--89.63 Acc-2 on \dsmosi{} and 0.404--0.422 MAE on \dssims{} at $\rho=0.4$). The number of combinations supervised in training grows as $J^M$ (1{,}728 for $J=12$ against 64 for $J=4$); we use $J=4$ in all other experiments.

\section{Qualitative Analysis}
\label{app:qual}

\begin{figure*}[t]
  \centering
  \includegraphics[width=\textwidth]{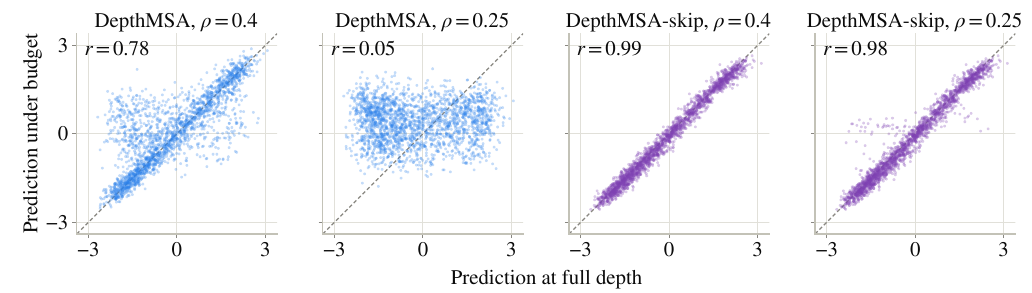}
  \caption{\textbf{Prediction under a budget against the prediction of the same model at full depth} on \dsmosi{}: one point per test sample and seed; $r$ is the Pearson correlation (mean of 3 seeds). With skipping, the predictions stay on the diagonal.}
  \label{fig:supp_drift}
\end{figure*}

\noindent\textbf{Prediction drift.} \Cref{fig:supp_drift} compares, for every \dsmosi{} test sample, the prediction under a budget with the prediction of the same model at full depth. With \method-skip, a tight budget changes the prediction only slightly: by 0.147 on average at $\rho=0.4$ and by 0.190 at $\rho=0.25$, and 96.7\% and 95.0\% of the predictions keep their sign. Without skipping, the predictions change by 0.580 and 1.363 on average. At $\rho=0.25$, where every modality runs only its first exit, they are almost unrelated to those at full depth ($r=0.05$) and concentrate around a small positive value, so only 50.3\% keep their sign. At $\rho=0.4$, the samples whose text keeps at least its third exit remain on the diagonal, and the others scatter.

\begin{figure*}[t]
  \centering
  \includegraphics[width=\textwidth]{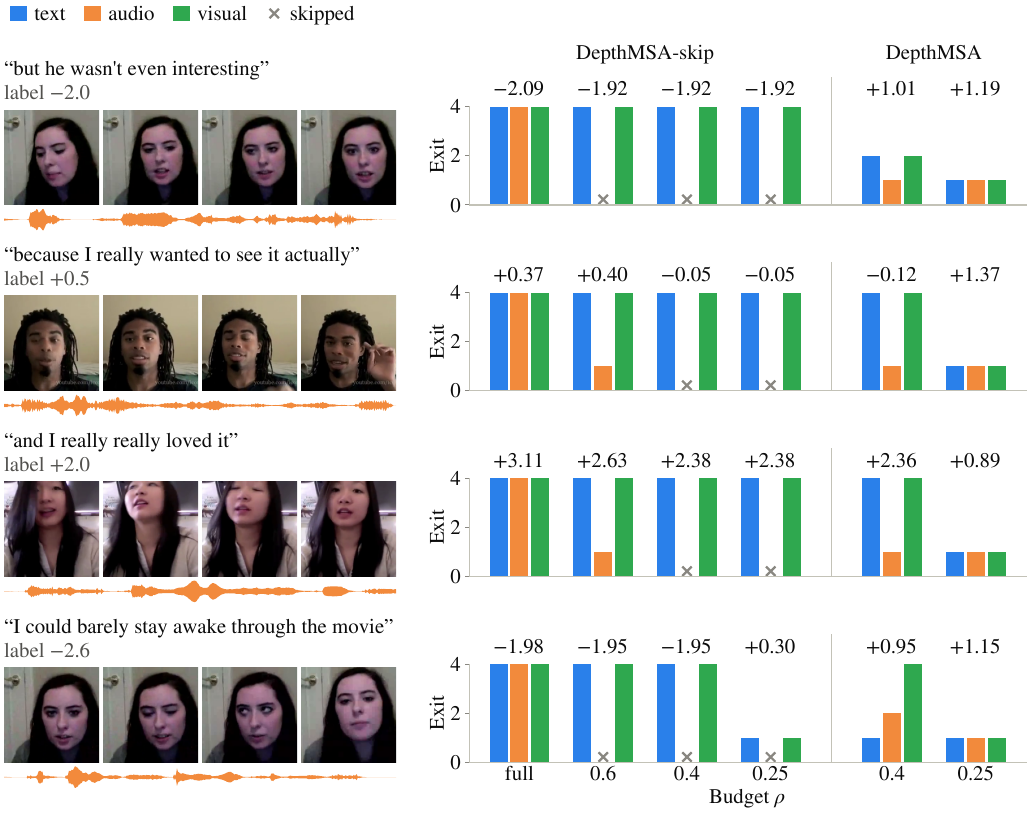}
  \caption{\textbf{Four utterances of \cref{tab:supp_cases}} with their frames, audio and transcript. Bars: exit selected for each modality under each budget; numbers: the resulting prediction.}
  \label{fig:supp_cases}
\end{figure*}

\noindent\textbf{Examples.} \Cref{tab:supp_cases} lists individual utterances. When the words carry the sentiment, \method-skip skips audio and returns nearly the same prediction at every budget, whereas \method{} without skipping must cut text to an early exit and can even flip the sign. Skipping audio is not free for every sample: it weakens a strongly negative prediction ($-2.52$ to $-2.10$) and flips a mildly positive one ($+0.40$ to $-0.05$), but it can also correct an overestimate ($+3.11$ to $+2.38$). Finally, the cap is per sample: in the last example, text accounts for a large share of the cost, so at $\rho=0.25$ even \method-skip must cut text to its first exit, and the prediction fails. \Cref{fig:supp_cases} shows four of these utterances with their frames and audio.

\begin{table*}[t]
\centering
\caption{\textbf{Full results on CMU-MOSI at every budget.} Mean $\pm$ std over $n$ seeds. GFLOPs and latency are measured end to end (batch size 1, A100). ``dense'' rows are separately trained models without exits. Budgets of \method{} and \method-skip are on different scales (\cref{sec:setup}); compare on GFLOPs. ADMN ``full depth (stage 1)'' is its LayerDrop-trained backbone without a controller.}
\label{tab:supp_main_mosi}
\fittable{\textwidth}{%
\begin{tabular}{lcccccccccc}
\toprule
Method & $n$ & Acc-2 & F1 & Acc-2 (has-0) & Acc-5 & Acc-7 & MAE & Corr & GFLOPs & Lat. (ms) \\
\midrule
dense TAV (no exits) & 3 & 88.77 {\scriptsize$\pm$} 1.53 & 88.79 {\scriptsize$\pm$} 1.51 & 86.78 {\scriptsize$\pm$} 1.35 & 56.22 {\scriptsize$\pm$} 2.04 & 47.81 {\scriptsize$\pm$} 1.83 & 0.595 {\scriptsize$\pm$} 0.028 & 0.868 {\scriptsize$\pm$} 0.015 & 139.2 {\scriptsize$\pm$} 0.0 & 56.8 {\scriptsize$\pm$} 0.4 \\
dense TV (no audio) & 3 & 89.12 {\scriptsize$\pm$} 0.64 & 89.14 {\scriptsize$\pm$} 0.62 & 87.07 {\scriptsize$\pm$} 0.55 & 58.89 {\scriptsize$\pm$} 0.76 & 50.63 {\scriptsize$\pm$} 0.99 & 0.575 {\scriptsize$\pm$} 0.006 & 0.875 {\scriptsize$\pm$} 0.002 & 27.0 {\scriptsize$\pm$} 0.0 & 33.7 {\scriptsize$\pm$} 0.5 \\
dense T (text only) & 3 & 89.33 {\scriptsize$\pm$} 0.85 & 89.34 {\scriptsize$\pm$} 0.84 & 87.12 {\scriptsize$\pm$} 0.89 & 57.58 {\scriptsize$\pm$} 1.54 & 49.66 {\scriptsize$\pm$} 1.54 & 0.587 {\scriptsize$\pm$} 0.031 & 0.872 {\scriptsize$\pm$} 0.010 & 26.9 {\scriptsize$\pm$} 0.0 & 31.0 {\scriptsize$\pm$} 0.3 \\
ADMN full depth (stage 1) & 3 & 88.82 {\scriptsize$\pm$} 0.69 & 88.84 {\scriptsize$\pm$} 0.67 & 87.03 {\scriptsize$\pm$} 0.67 & 55.49 {\scriptsize$\pm$} 3.94 & 47.18 {\scriptsize$\pm$} 2.32 & 0.643 {\scriptsize$\pm$} 0.037 & 0.854 {\scriptsize$\pm$} 0.007 & 139.2 {\scriptsize$\pm$} 0.0 & 56.2 {\scriptsize$\pm$} 1.8 \\
ADMN $\rho{=}0.8$ & 3 & 88.92 {\scriptsize$\pm$} 1.22 & 88.92 {\scriptsize$\pm$} 1.24 & 87.03 {\scriptsize$\pm$} 1.39 & 55.59 {\scriptsize$\pm$} 4.16 & 47.72 {\scriptsize$\pm$} 2.93 & 0.644 {\scriptsize$\pm$} 0.051 & 0.849 {\scriptsize$\pm$} 0.017 & 118.7 {\scriptsize$\pm$} 1.1 & 66.0 {\scriptsize$\pm$} 2.1 \\
\bottomrule
\end{tabular}}
\end{table*}

\begin{table*}[t]
\centering
\caption{\textbf{Full results on CMU-MOSEI at every budget.} Mean $\pm$ std over $n$ seeds. GFLOPs and latency are measured end to end (batch size 1, A100). ``dense'' rows are separately trained models without exits. Budgets of \method{} and \method-skip are on different scales (\cref{sec:setup}); compare on GFLOPs. ADMN ``full depth (stage 1)'' is its LayerDrop-trained backbone without a controller.}
\label{tab:supp_main_mosei}
\fittable{\textwidth}{%
\begin{tabular}{lcccccccccc}
\toprule
Method & $n$ & Acc-2 & F1 & Acc-2 (has-0) & Acc-5 & Acc-7 & MAE & Corr & GFLOPs & Lat. (ms) \\
\midrule
dense TAV (no exits) & 3 & 87.91 {\scriptsize$\pm$} 0.46 & 87.95 {\scriptsize$\pm$} 0.45 & 82.89 {\scriptsize$\pm$} 0.75 & 58.44 {\scriptsize$\pm$} 0.73 & 56.48 {\scriptsize$\pm$} 0.73 & 0.488 {\scriptsize$\pm$} 0.010 & 0.818 {\scriptsize$\pm$} 0.004 & 172.9 {\scriptsize$\pm$} 0.0 & 56.5 {\scriptsize$\pm$} 0.3 \\
dense TV (no audio) & 3 & 88.37 {\scriptsize$\pm$} 0.22 & 88.37 {\scriptsize$\pm$} 0.20 & 83.84 {\scriptsize$\pm$} 1.20 & 57.32 {\scriptsize$\pm$} 0.60 & 55.63 {\scriptsize$\pm$} 0.43 & 0.493 {\scriptsize$\pm$} 0.004 & 0.809 {\scriptsize$\pm$} 0.005 & 36.4 {\scriptsize$\pm$} 0.0 & 34.0 {\scriptsize$\pm$} 0.4 \\
dense T (text only) & 3 & 86.66 {\scriptsize$\pm$} 0.42 & 86.77 {\scriptsize$\pm$} 0.33 & 80.15 {\scriptsize$\pm$} 1.68 & 57.93 {\scriptsize$\pm$} 0.09 & 55.96 {\scriptsize$\pm$} 0.08 & 0.492 {\scriptsize$\pm$} 0.005 & 0.810 {\scriptsize$\pm$} 0.001 & 36.2 {\scriptsize$\pm$} 0.0 & 31.4 {\scriptsize$\pm$} 0.5 \\
ADMN full depth (stage 1) & 3 & 87.31 {\scriptsize$\pm$} 0.32 & 87.43 {\scriptsize$\pm$} 0.29 & 80.69 {\scriptsize$\pm$} 0.95 & 58.38 {\scriptsize$\pm$} 0.68 & 56.46 {\scriptsize$\pm$} 0.58 & 0.487 {\scriptsize$\pm$} 0.000 & 0.817 {\scriptsize$\pm$} 0.003 & 172.9 {\scriptsize$\pm$} 0.0 & 55.2 {\scriptsize$\pm$} 0.6 \\
ADMN $\rho{=}0.8$ & 3 & 87.56 {\scriptsize$\pm$} 1.06 & 87.64 {\scriptsize$\pm$} 0.99 & 81.63 {\scriptsize$\pm$} 2.52 & 57.86 {\scriptsize$\pm$} 0.79 & 55.93 {\scriptsize$\pm$} 0.72 & 0.493 {\scriptsize$\pm$} 0.009 & 0.816 {\scriptsize$\pm$} 0.005 & 144.8 {\scriptsize$\pm$} 0.1 & 66.1 {\scriptsize$\pm$} 0.6 \\
\bottomrule
\end{tabular}}
\end{table*}

\begin{table*}[t]
\centering
\caption{\textbf{Full results on CH-SIMS at every budget.} Mean $\pm$ std over $n$ seeds. GFLOPs and latency are measured end to end (batch size 1, A100). ``dense'' rows are separately trained models without exits. Budgets of \method{} and \method-skip are on different scales (\cref{sec:setup}); compare on GFLOPs. ADMN ``full depth (stage 1)'' is its LayerDrop-trained backbone without a controller.}
\label{tab:supp_main_sims}
\fittable{\textwidth}{%
\begin{tabular}{lccccccccc}
\toprule
Method & $n$ & Acc-2 & F1 & Acc-3 & Acc-5 & MAE & Corr & GFLOPs & Lat. (ms) \\
\midrule
dense TAV (no exits) & 3 & 79.43 {\scriptsize$\pm$} 0.79 & 79.72 {\scriptsize$\pm$} 0.60 & 69.51 {\scriptsize$\pm$} 0.33 & 47.26 {\scriptsize$\pm$} 1.58 & 0.377 {\scriptsize$\pm$} 0.011 & 0.661 {\scriptsize$\pm$} 0.013 & 55.2 {\scriptsize$\pm$} 0.0 & 21.2 {\scriptsize$\pm$} 0.2 \\
dense TV (no audio) & 3 & 78.77 {\scriptsize$\pm$} 1.33 & 78.86 {\scriptsize$\pm$} 1.37 & 66.66 {\scriptsize$\pm$} 2.41 & 42.67 {\scriptsize$\pm$} 3.50 & 0.409 {\scriptsize$\pm$} 0.015 & 0.613 {\scriptsize$\pm$} 0.018 & 3.1 {\scriptsize$\pm$} 0.0 & 11.4 {\scriptsize$\pm$} 0.2 \\
dense T (text only) & 3 & 78.19 {\scriptsize$\pm$} 1.54 & 78.12 {\scriptsize$\pm$} 1.15 & 67.18 {\scriptsize$\pm$} 0.66 & 43.40 {\scriptsize$\pm$} 1.49 & 0.411 {\scriptsize$\pm$} 0.007 & 0.610 {\scriptsize$\pm$} 0.009 & 3.1 {\scriptsize$\pm$} 0.0 & 8.6 {\scriptsize$\pm$} 0.1 \\
ADMN full depth (stage 1) & 3 & 80.38 {\scriptsize$\pm$} 0.63 & 80.79 {\scriptsize$\pm$} 0.56 & 69.66 {\scriptsize$\pm$} 0.83 & 46.53 {\scriptsize$\pm$} 1.21 & 0.361 {\scriptsize$\pm$} 0.002 & 0.715 {\scriptsize$\pm$} 0.006 & 55.2 {\scriptsize$\pm$} 0.0 & 19.1 {\scriptsize$\pm$} 0.1 \\
ADMN $\rho{=}0.8$ & 3 & 79.87 {\scriptsize$\pm$} 1.16 & 80.29 {\scriptsize$\pm$} 1.15 & 69.73 {\scriptsize$\pm$} 1.10 & 46.02 {\scriptsize$\pm$} 1.91 & 0.362 {\scriptsize$\pm$} 0.005 & 0.711 {\scriptsize$\pm$} 0.005 & 47.2 {\scriptsize$\pm$} 0.0 & 27.8 {\scriptsize$\pm$} 0.2 \\
ADMN $\rho{=}0.6$ & 3 & 80.89 {\scriptsize$\pm$} 0.55 & 81.12 {\scriptsize$\pm$} 0.58 & 69.07 {\scriptsize$\pm$} 1.65 & 45.59 {\scriptsize$\pm$} 1.96 & 0.373 {\scriptsize$\pm$} 0.005 & 0.688 {\scriptsize$\pm$} 0.011 & 39.7 {\scriptsize$\pm$} 0.0 & 25.9 {\scriptsize$\pm$} 0.2 \\
\bottomrule
\end{tabular}}
\end{table*}

\begin{table*}[t]
\centering
\caption{\textbf{Allocation rules on CMU-MOSI} (same trained models, mean of 3 seeds). Each cell: MAE / Acc-2 (fraction of the full encoder FLOPs actually used). The oracle picks the best feasible combination using the test labels and is an optimistic reference, not a method.}
\label{tab:supp_alloc_mosi_noskip}
\fittable{\textwidth}{%
\begin{tabular}{lcccccc}
\toprule
$\rho$ & \method{} & learned importance & text-first & uniform & random & oracle \\
\midrule
1.0 & 0.615 / 88.7 (0.70) & 0.606 / 88.6 (1.00) & 0.606 / 88.6 (1.00) & 0.606 / 88.6 (1.00) & 0.606 / 88.6 (1.00) & 0.309 / 96.4 (0.67) \\
0.9 & 0.619 / 88.8 (0.52) & 0.621 / 88.7 (0.82) & 0.621 / 88.7 (0.82) & 0.663 / 87.5 (0.75) & 0.882 / 79.2 (0.84) & 0.327 / 96.2 (0.59) \\
0.8 & 0.619 / 88.8 (0.52) & 0.623 / 88.6 (0.67) & 0.623 / 88.6 (0.67) & 0.663 / 87.5 (0.75) & 0.828 / 80.0 (0.71) & 0.340 / 95.8 (0.54) \\
0.7 & 0.620 / 88.8 (0.52) & 0.621 / 88.4 (0.63) & 0.621 / 88.4 (0.63) & 1.376 / 58.2 (0.50) & 0.971 / 74.9 (0.65) & 0.354 / 95.4 (0.50) \\
0.6 & 0.622 / 88.7 (0.47) & 0.621 / 88.7 (0.48) & 0.621 / 88.7 (0.48) & 1.376 / 58.2 (0.50) & 0.797 / 81.1 (0.53) & 0.376 / 94.8 (0.46) \\
0.5 & 0.629 / 88.9 (0.44) & 0.628 / 88.8 (0.44) & 0.628 / 88.8 (0.44) & 1.376 / 58.2 (0.50) & 1.031 / 71.4 (0.44) & 0.418 / 94.4 (0.41) \\
0.4 & 0.938 / 75.4 (0.35) & 0.945 / 75.0 (0.36) & 0.945 / 75.0 (0.36) & 1.558 / 49.6 (0.25) & 0.948 / 74.9 (0.36) & 0.792 / 79.6 (0.33) \\
0.3 & 1.521 / 50.8 (0.25) & 1.517 / 50.3 (0.25) & 1.517 / 50.3 (0.25) & 1.558 / 49.6 (0.25) & 1.518 / 50.3 (0.25) & 1.468 / 52.5 (0.25) \\
0.25 & 1.558 / 49.6 (0.25) & 1.558 / 49.6 (0.25) & 1.558 / 49.6 (0.25) & 1.558 / 49.6 (0.25) & 1.558 / 49.6 (0.25) & 1.558 / 49.6 (0.25) \\
0.2 & -- & -- & -- & -- & -- & -- \\
0.1 & -- & -- & -- & -- & -- & -- \\
\bottomrule
\end{tabular}}
\end{table*}

\begin{table*}[t]
\centering
\caption{\textbf{Allocation rules on CMU-MOSI with modality skipping} (same trained models, mean of 3 seeds). Each cell: MAE / Acc-2 (fraction of the full encoder FLOPs actually used). The oracle picks the best feasible combination using the test labels and is an optimistic reference, not a method.}
\label{tab:supp_alloc_mosi_skip}
\fittable{\textwidth}{%
\begin{tabular}{lccccc}
\toprule
$\rho$ & \method{} & text-first & uniform & random & oracle \\
\midrule
1.0 & 0.614 / 88.0 (0.45) & 0.619 / 88.0 (1.00) & 0.619 / 88.0 (1.00) & 0.619 / 88.0 (1.00) & 0.212 / 98.9 (0.56) \\
0.9 & 0.622 / 88.0 (0.33) & 0.650 / 88.0 (0.87) & 0.682 / 88.0 (0.82) & 0.921 / 77.7 (0.87) & 0.219 / 98.9 (0.50) \\
0.8 & 0.622 / 88.0 (0.33) & 0.654 / 88.0 (0.75) & 1.345 / 61.1 (0.65) & 0.971 / 75.4 (0.76) & 0.233 / 98.8 (0.44) \\
0.7 & 0.622 / 88.0 (0.33) & 0.637 / 88.0 (0.62) & 1.345 / 61.1 (0.65) & 0.930 / 75.9 (0.65) & 0.247 / 98.7 (0.38) \\
0.6 & 0.614 / 88.1 (0.21) & 0.623 / 88.5 (0.28) & 1.502 / 52.7 (0.47) & 0.990 / 73.2 (0.43) & 0.267 / 98.7 (0.30) \\
0.5 & 0.610 / 88.1 (0.19) & 0.612 / 88.4 (0.20) & 1.486 / 52.7 (0.45) & 1.035 / 70.4 (0.33) & 0.309 / 98.7 (0.19) \\
0.4 & 0.610 / 88.1 (0.19) & 0.612 / 88.4 (0.20) & 1.424 / 42.2 (0.00) & 0.687 / 84.8 (0.21) & 0.335 / 97.4 (0.14) \\
0.3 & 0.610 / 88.0 (0.18) & 0.613 / 88.4 (0.20) & 1.424 / 42.2 (0.00) & 0.615 / 88.3 (0.20) & 0.342 / 97.2 (0.14) \\
0.25 & 0.637 / 86.4 (0.17) & 0.643 / 86.6 (0.19) & 1.424 / 42.2 (0.00) & 0.643 / 86.6 (0.19) & 0.368 / 96.2 (0.13) \\
0.2 & 0.693 / 84.7 (0.15) & 0.703 / 84.5 (0.16) & 1.424 / 42.2 (0.00) & 0.703 / 84.5 (0.16) & 0.431 / 95.1 (0.12) \\
0.1 & 1.239 / 59.5 (0.06) & 1.255 / 58.6 (0.07) & 1.424 / 42.2 (0.00) & 1.257 / 58.4 (0.07) & 1.015 / 78.8 (0.05) \\
\bottomrule
\end{tabular}}
\end{table*}

\begin{table*}[t]
\centering
\caption{\textbf{Allocation rules on CMU-MOSEI} (same trained models, mean of 3 seeds). Each cell: MAE / Acc-2 (fraction of the full encoder FLOPs actually used). The oracle picks the best feasible combination using the test labels and is an optimistic reference, not a method.}
\label{tab:supp_alloc_mosei_noskip}
\fittable{\textwidth}{%
\begin{tabular}{lcccccc}
\toprule
$\rho$ & \method{} & learned importance & text-first & uniform & random & oracle \\
\midrule
1.0 & 0.495 / 87.1 (0.94) & 0.495 / 87.1 (1.00) & 0.495 / 87.1 (1.00) & 0.495 / 87.1 (1.00) & 0.495 / 87.1 (1.00) & 0.251 / 95.5 (0.59) \\
0.9 & 0.496 / 87.1 (0.82) & 0.496 / 87.2 (0.82) & 0.496 / 87.2 (0.82) & 0.495 / 87.6 (0.75) & 0.499 / 86.8 (0.84) & 0.253 / 95.4 (0.55) \\
0.8 & 0.496 / 87.1 (0.70) & 0.498 / 87.1 (0.68) & 0.498 / 87.1 (0.68) & 0.495 / 87.6 (0.75) & 0.514 / 86.3 (0.72) & 0.257 / 95.3 (0.51) \\
0.7 & 0.498 / 87.1 (0.63) & 0.498 / 87.1 (0.63) & 0.498 / 87.1 (0.63) & 0.505 / 87.1 (0.50) & 0.503 / 86.7 (0.65) & 0.262 / 95.2 (0.47) \\
0.6 & 0.499 / 87.1 (0.50) & 0.499 / 87.1 (0.50) & 0.499 / 87.1 (0.50) & 0.505 / 87.1 (0.50) & 0.513 / 86.5 (0.53) & 0.270 / 95.0 (0.42) \\
0.5 & 0.498 / 87.2 (0.44) & 0.498 / 87.2 (0.44) & 0.498 / 87.2 (0.44) & 0.505 / 87.1 (0.50) & 0.517 / 86.2 (0.45) & 0.282 / 94.7 (0.38) \\
0.4 & 0.505 / 86.9 (0.36) & 0.505 / 87.0 (0.36) & 0.505 / 87.0 (0.36) & 0.566 / 84.5 (0.25) & 0.509 / 87.0 (0.36) & 0.335 / 93.3 (0.31) \\
0.3 & 0.561 / 84.5 (0.26) & 0.561 / 84.7 (0.26) & 0.561 / 84.7 (0.26) & 0.566 / 84.5 (0.25) & 0.561 / 84.6 (0.26) & 0.522 / 86.4 (0.25) \\
0.25 & 0.566 / 84.5 (0.25) & 0.566 / 84.5 (0.25) & 0.566 / 84.5 (0.25) & 0.566 / 84.5 (0.25) & 0.566 / 84.5 (0.25) & 0.566 / 84.5 (0.25) \\
0.2 & -- & -- & -- & -- & -- & -- \\
0.1 & -- & -- & -- & -- & -- & -- \\
\bottomrule
\end{tabular}}
\end{table*}

\begin{table*}[t]
\centering
\caption{\textbf{Allocation rules on CMU-MOSEI with modality skipping} (same trained models, mean of 3 seeds). Each cell: MAE / Acc-2 (fraction of the full encoder FLOPs actually used). The oracle picks the best feasible combination using the test labels and is an optimistic reference, not a method.}
\label{tab:supp_alloc_mosei_skip}
\fittable{\textwidth}{%
\begin{tabular}{lccccc}
\toprule
$\rho$ & \method{} & text-first & uniform & random & oracle \\
\midrule
1.0 & 0.494 / 86.6 (0.92) & 0.495 / 86.3 (1.00) & 0.495 / 86.3 (1.00) & 0.495 / 86.3 (1.00) & 0.153 / 99.0 (0.49) \\
0.9 & 0.494 / 87.1 (0.84) & 0.497 / 87.0 (0.87) & 0.494 / 87.2 (0.82) & 0.515 / 85.7 (0.87) & 0.154 / 99.0 (0.46) \\
0.8 & 0.497 / 86.6 (0.73) & 0.499 / 86.8 (0.75) & 0.512 / 85.6 (0.65) & 0.539 / 85.0 (0.76) & 0.158 / 99.0 (0.42) \\
0.7 & 0.498 / 86.8 (0.67) & 0.501 / 86.7 (0.62) & 0.512 / 85.6 (0.65) & 0.555 / 83.9 (0.65) & 0.164 / 98.8 (0.36) \\
0.6 & 0.499 / 86.8 (0.57) & 0.505 / 87.0 (0.30) & 0.554 / 84.3 (0.47) & 0.563 / 84.2 (0.44) & 0.174 / 98.5 (0.29) \\
0.5 & 0.504 / 87.2 (0.17) & 0.506 / 86.9 (0.20) & 0.563 / 83.5 (0.42) & 0.550 / 84.5 (0.34) & 0.191 / 98.2 (0.19) \\
0.4 & 0.504 / 87.2 (0.17) & 0.506 / 86.9 (0.20) & 0.841 / 62.8 (0.00) & 0.552 / 84.6 (0.21) & 0.221 / 97.5 (0.10) \\
0.3 & 0.504 / 87.3 (0.17) & 0.506 / 86.9 (0.20) & 0.841 / 62.8 (0.00) & 0.506 / 86.9 (0.20) & 0.230 / 97.3 (0.09) \\
0.25 & 0.505 / 87.2 (0.16) & 0.506 / 86.9 (0.18) & 0.841 / 62.8 (0.00) & 0.506 / 86.9 (0.18) & 0.233 / 97.2 (0.09) \\
0.2 & 0.508 / 87.0 (0.15) & 0.508 / 86.6 (0.16) & 0.841 / 62.8 (0.00) & 0.508 / 86.6 (0.16) & 0.242 / 97.1 (0.08) \\
0.1 & 0.542 / 85.3 (0.07) & 0.541 / 85.2 (0.07) & 0.841 / 62.8 (0.00) & 0.541 / 85.2 (0.07) & 0.325 / 95.2 (0.04) \\
\bottomrule
\end{tabular}}
\end{table*}

\begin{table*}[t]
\centering
\caption{\textbf{Allocation rules on CH-SIMS} (same trained models, mean of 3 seeds). Each cell: MAE / Acc-2 (fraction of the full encoder FLOPs actually used). The oracle picks the best feasible combination using the test labels and is an optimistic reference, not a method.}
\label{tab:supp_alloc_sims_noskip}
\fittable{\textwidth}{%
\begin{tabular}{lcccccc}
\toprule
$\rho$ & \method{} & learned importance & text-first & uniform & random & oracle \\
\midrule
1.0 & 0.360 / 79.1 (0.83) & 0.386 / 78.7 (1.00) & 0.386 / 78.7 (1.00) & 0.386 / 78.7 (1.00) & 0.386 / 78.7 (1.00) & 0.130 / 90.0 (0.58) \\
0.9 & 0.360 / 79.1 (0.75) & 0.380 / 79.7 (0.77) & 0.380 / 79.7 (0.77) & 0.362 / 78.3 (0.75) & 0.383 / 79.3 (0.78) & 0.134 / 89.9 (0.52) \\
0.8 & 0.360 / 79.1 (0.75) & 0.380 / 79.7 (0.77) & 0.380 / 79.7 (0.77) & 0.362 / 78.3 (0.75) & 0.380 / 79.7 (0.77) & 0.134 / 89.9 (0.52) \\
0.7 & 0.376 / 77.7 (0.53) & 0.390 / 77.5 (0.55) & 0.390 / 77.5 (0.55) & 0.470 / 69.8 (0.50) & 0.403 / 77.1 (0.57) & 0.163 / 89.1 (0.43) \\
0.6 & 0.376 / 77.7 (0.53) & 0.390 / 77.5 (0.55) & 0.390 / 77.5 (0.55) & 0.470 / 69.8 (0.50) & 0.390 / 77.5 (0.55) & 0.175 / 88.6 (0.41) \\
0.5 & 0.397 / 77.5 (0.32) & 0.398 / 77.0 (0.32) & 0.398 / 77.0 (0.32) & 0.470 / 69.8 (0.50) & 0.428 / 74.3 (0.40) & 0.190 / 88.8 (0.35) \\
0.4 & 0.397 / 77.5 (0.32) & 0.398 / 77.0 (0.32) & 0.398 / 77.0 (0.32) & 0.502 / 68.5 (0.25) & 0.398 / 77.0 (0.32) & 0.227 / 85.9 (0.30) \\
0.3 & 0.446 / 73.2 (0.28) & 0.450 / 72.9 (0.29) & 0.450 / 72.9 (0.29) & 0.502 / 68.5 (0.25) & 0.452 / 71.9 (0.29) & 0.365 / 77.9 (0.27) \\
0.25 & 0.502 / 68.5 (0.25) & 0.502 / 68.5 (0.25) & 0.502 / 68.5 (0.25) & 0.502 / 68.5 (0.25) & 0.502 / 68.5 (0.25) & 0.502 / 68.5 (0.25) \\
0.2 & -- & -- & -- & -- & -- & -- \\
0.1 & -- & -- & -- & -- & -- & -- \\
\bottomrule
\end{tabular}}
\end{table*}

\begin{table*}[t]
\centering
\caption{\textbf{Allocation rules on CH-SIMS with modality skipping} (same trained models, mean of 3 seeds). Each cell: MAE / Acc-2 (fraction of the full encoder FLOPs actually used). The oracle picks the best feasible combination using the test labels and is an optimistic reference, not a method.}
\label{tab:supp_alloc_sims_skip}
\fittable{\textwidth}{%
\begin{tabular}{lccccc}
\toprule
$\rho$ & \method{} & text-first & uniform & random & oracle \\
\midrule
1.0 & 0.351 / 77.4 (0.98) & 0.384 / 78.7 (1.00) & 0.384 / 78.7 (1.00) & 0.384 / 78.7 (1.00) & 0.102 / 91.2 (0.55) \\
0.9 & 0.352 / 77.9 (0.84) & 0.374 / 79.4 (0.85) & 0.352 / 77.6 (0.84) & 0.373 / 79.4 (0.85) & 0.105 / 91.0 (0.50) \\
0.8 & 0.371 / 79.2 (0.70) & 0.383 / 78.3 (0.71) & 0.460 / 70.4 (0.68) & 0.409 / 77.0 (0.73) & 0.131 / 90.6 (0.43) \\
0.7 & 0.381 / 77.9 (0.61) & 0.394 / 76.4 (0.57) & 0.460 / 70.4 (0.68) & 0.391 / 77.0 (0.63) & 0.143 / 90.8 (0.38) \\
0.6 & 0.394 / 78.0 (0.56) & 0.395 / 76.2 (0.56) & 0.510 / 64.5 (0.52) & 0.395 / 76.2 (0.56) & 0.173 / 89.9 (0.28) \\
0.5 & 0.409 / 77.0 (0.06) & 0.410 / 76.4 (0.06) & 0.601 / 69.4 (0.00) & 0.434 / 75.0 (0.12) & 0.228 / 88.9 (0.07) \\
0.4 & 0.409 / 77.0 (0.06) & 0.410 / 76.4 (0.06) & 0.601 / 69.4 (0.00) & 0.410 / 76.4 (0.06) & 0.234 / 89.2 (0.04) \\
0.3 & 0.409 / 77.0 (0.06) & 0.410 / 76.4 (0.06) & 0.601 / 69.4 (0.00) & 0.410 / 76.4 (0.06) & 0.234 / 89.2 (0.04) \\
0.25 & 0.409 / 77.0 (0.06) & 0.410 / 76.4 (0.06) & 0.601 / 69.4 (0.00) & 0.410 / 76.4 (0.06) & 0.234 / 89.2 (0.04) \\
0.2 & 0.409 / 77.0 (0.06) & 0.410 / 76.4 (0.06) & 0.601 / 69.4 (0.00) & 0.410 / 76.4 (0.06) & 0.234 / 89.2 (0.04) \\
0.1 & 0.408 / 77.0 (0.06) & 0.409 / 76.4 (0.06) & 0.601 / 69.4 (0.00) & 0.409 / 76.4 (0.06) & 0.235 / 89.2 (0.04) \\
\bottomrule
\end{tabular}}
\end{table*}

\begin{table*}[t]
\centering
\caption{\textbf{Training ablation on CMU-MOSI}.}
\label{tab:supp_abl_mosi}
\fittable{\textwidth}{%
\begin{tabular}{lccccccccc}
\toprule
Variant & Acc-2 & F1 & Acc-2 (has-0) & Acc-5 & Acc-7 & MAE & Corr & GFLOPs & Lat. (ms) \\
\midrule
ours full & 88.41 & 88.44 & 86.15 & 54.52 & 46.50 & 0.621 & 0.864 & 139.2 & 61.8 \\
ours $\rho{=}0.6$ & 88.57 & 88.56 & 86.15 & 56.56 & 48.25 & 0.625 & 0.857 & 86.3 & 49.9 \\
ours $\rho{=}0.4$ & 73.63 & 73.78 & 72.45 & 39.94 & 34.11 & 0.971 & 0.626 & 77.6 & 38.5 \\
ours $\rho{=}0.3$ & 53.51 & 53.06 & 54.23 & 18.66 & 17.93 & 1.459 & 0.071 & 66.6 & 25.0 \\
ours $\rho{=}0.25$ & 53.20 & 52.60 & 53.79 & 18.51 & 18.51 & 1.486 & 0.004 & 66.2 & 23.6 \\
w/o combinatorial training full & 89.94 & 89.90 & 87.61 & 56.71 & 48.83 & 0.586 & 0.868 & 139.2 & 63.4 \\
w/o combinatorial training $\rho{=}0.6$ & 88.41 & 88.43 & 86.44 & 54.96 & 46.94 & 0.626 & 0.856 & 87.9 & 51.3 \\
\bottomrule
\end{tabular}}
\end{table*}

\begin{table*}[t]
\centering
\caption{\textbf{Training ablation on CMU-MOSEI}.}
\label{tab:supp_abl_mosei}
\fittable{\textwidth}{%
\begin{tabular}{lccccccccc}
\toprule
Variant & Acc-2 & F1 & Acc-2 (has-0) & Acc-5 & Acc-7 & MAE & Corr & GFLOPs & Lat. (ms) \\
\midrule
ours full & 87.75 & 87.69 & 85.02 & 57.82 & 55.72 & 0.494 & 0.812 & 172.9 & 63.7 \\
ours $\rho{=}0.6$ & 87.62 & 87.51 & 85.30 & 57.48 & 55.20 & 0.500 & 0.811 & 109.7 & 50.8 \\
ours $\rho{=}0.4$ & 87.81 & 87.78 & 84.42 & 56.75 & 54.75 & 0.503 & 0.807 & 94.9 & 42.9 \\
ours $\rho{=}0.3$ & 84.84 & 84.87 & 80.40 & 53.55 & 51.88 & 0.550 & 0.762 & 81.7 & 26.8 \\
ours $\rho{=}0.25$ & 84.62 & 84.65 & 80.04 & 53.38 & 51.62 & 0.554 & 0.758 & 81.3 & 24.4 \\
w/o combinatorial training full & 86.08 & 86.20 & 79.63 & 56.75 & 54.73 & 0.501 & 0.805 & 172.9 & 61.9 \\
w/o combinatorial training $\rho{=}0.6$ & 84.26 & 84.47 & 76.86 & 56.17 & 54.30 & 0.518 & 0.791 & 109.7 & 50.2 \\
\bottomrule
\end{tabular}}
\end{table*}

\begin{table*}[t]
\centering
\caption{\textbf{Training ablation on CH-SIMS}.}
\label{tab:supp_abl_sims}
\fittable{\textwidth}{%
\begin{tabular}{lcccccccc}
\toprule
Variant & Acc-2 & F1 & Acc-3 & Acc-5 & MAE & Corr & GFLOPs & Lat. (ms) \\
\midrule
ours full & 79.43 & 79.96 & 68.93 & 47.48 & 0.378 & 0.675 & 55.2 & 24.7 \\
ours $\rho{=}0.6$ & 77.90 & 78.59 & 71.33 & 43.98 & 0.365 & 0.721 & 38.3 & 18.6 \\
ours $\rho{=}0.4$ & 77.02 & 77.52 & 68.71 & 45.30 & 0.392 & 0.637 & 30.9 & 21.6 \\
ours $\rho{=}0.3$ & 73.52 & 74.13 & 63.68 & 36.98 & 0.437 & 0.565 & 29.7 & 14.2 \\
ours $\rho{=}0.25$ & 68.49 & 68.43 & 55.80 & 28.67 & 0.504 & 0.406 & 28.6 & 10.4 \\
w/o combinatorial training full & 77.02 & 77.79 & 68.71 & 46.83 & 0.374 & 0.670 & 55.2 & 24.2 \\
w/o combinatorial training $\rho{=}0.6$ & 80.53 & 80.69 & 67.61 & 44.20 & 0.395 & 0.644 & 39.0 & 19.8 \\
\bottomrule
\end{tabular}}
\end{table*}

\begin{table*}[t]
\centering
\caption{\textbf{Backbones on CMU-MOSI} ($n$ = number of seeds).}
\label{tab:supp_bb_mosi}
\fittable{\textwidth}{%
\begin{tabular}{lcccccccccc}
\toprule
Backbone & $n$ & Acc-2 & F1 & Acc-2 (has-0) & Acc-5 & Acc-7 & MAE & Corr & GFLOPs & Lat. (ms) \\
\midrule
RoBERTa-L + data2vec + Facet full & 3 & 88.62 {\scriptsize$\pm$} 1.23 & 88.62 {\scriptsize$\pm$} 1.18 & 86.44 {\scriptsize$\pm$} 1.05 & 56.27 {\scriptsize$\pm$} 1.54 & 48.15 {\scriptsize$\pm$} 1.44 & 0.606 {\scriptsize$\pm$} 0.016 & 0.860 {\scriptsize$\pm$} 0.003 & 139.2 {\scriptsize$\pm$} 0.0 & 61.6 {\scriptsize$\pm$} 0.6 \\
RoBERTa-L + data2vec + Facet $\rho{=}0.8$ & 3 & 88.77 {\scriptsize$\pm$} 1.23 & 88.77 {\scriptsize$\pm$} 1.19 & 86.39 {\scriptsize$\pm$} 0.97 & 56.37 {\scriptsize$\pm$} 0.33 & 47.96 {\scriptsize$\pm$} 0.39 & 0.619 {\scriptsize$\pm$} 0.007 & 0.853 {\scriptsize$\pm$} 0.003 & 92.3 {\scriptsize$\pm$} 10.1 & 52.6 {\scriptsize$\pm$} 2.9 \\
RoBERTa-L + data2vec + Facet $\rho{=}0.6$ & 3 & 88.67 {\scriptsize$\pm$} 1.07 & 88.67 {\scriptsize$\pm$} 1.03 & 86.30 {\scriptsize$\pm$} 0.81 & 56.12 {\scriptsize$\pm$} 0.39 & 47.57 {\scriptsize$\pm$} 0.66 & 0.622 {\scriptsize$\pm$} 0.003 & 0.852 {\scriptsize$\pm$} 0.005 & 86.9 {\scriptsize$\pm$} 0.9 & 49.0 {\scriptsize$\pm$} 0.8 \\
RoBERTa-L + data2vec + Facet $\rho{=}0.5$ & 3 & 88.87 {\scriptsize$\pm$} 1.45 & 88.88 {\scriptsize$\pm$} 1.42 & 86.39 {\scriptsize$\pm$} 1.24 & 55.88 {\scriptsize$\pm$} 0.17 & 47.33 {\scriptsize$\pm$} 0.22 & 0.629 {\scriptsize$\pm$} 0.007 & 0.847 {\scriptsize$\pm$} 0.002 & 84.9 {\scriptsize$\pm$} 0.0 & 47.6 {\scriptsize$\pm$} 0.9 \\
RoBERTa-L + data2vec + Facet $\rho{=}0.4$ & 3 & 75.41 {\scriptsize$\pm$} 4.18 & 75.54 {\scriptsize$\pm$} 4.17 & 74.34 {\scriptsize$\pm$} 3.67 & 42.52 {\scriptsize$\pm$} 4.33 & 36.35 {\scriptsize$\pm$} 3.50 & 0.938 {\scriptsize$\pm$} 0.074 & 0.657 {\scriptsize$\pm$} 0.070 & 76.9 {\scriptsize$\pm$} 1.2 & 38.3 {\scriptsize$\pm$} 1.4 \\
\bottomrule
\end{tabular}}
\end{table*}

\begin{table*}[t]
\centering
\caption{\textbf{Backbones on CMU-MOSEI} ($n$ = number of seeds).}
\label{tab:supp_bb_mosei}
\fittable{\textwidth}{%
\begin{tabular}{lcccccccccc}
\toprule
Backbone & $n$ & Acc-2 & F1 & Acc-2 (has-0) & Acc-5 & Acc-7 & MAE & Corr & GFLOPs & Lat. (ms) \\
\midrule
RoBERTa-L + data2vec + Facet full & 3 & 87.14 {\scriptsize$\pm$} 1.06 & 87.13 {\scriptsize$\pm$} 0.97 & 82.95 {\scriptsize$\pm$} 2.95 & 57.55 {\scriptsize$\pm$} 0.27 & 55.71 {\scriptsize$\pm$} 0.05 & 0.495 {\scriptsize$\pm$} 0.001 & 0.808 {\scriptsize$\pm$} 0.003 & 172.9 {\scriptsize$\pm$} 0.0 & 62.5 {\scriptsize$\pm$} 1.1 \\
RoBERTa-L + data2vec + Facet $\rho{=}0.8$ & 3 & 87.06 {\scriptsize$\pm$} 0.94 & 87.06 {\scriptsize$\pm$} 0.88 & 82.41 {\scriptsize$\pm$} 2.38 & 57.64 {\scriptsize$\pm$} 0.64 & 55.63 {\scriptsize$\pm$} 0.55 & 0.496 {\scriptsize$\pm$} 0.002 & 0.808 {\scriptsize$\pm$} 0.005 & 135.2 {\scriptsize$\pm$} 6.5 & 57.5 {\scriptsize$\pm$} 0.2 \\
RoBERTa-L + data2vec + Facet $\rho{=}0.6$ & 3 & 87.07 {\scriptsize$\pm$} 0.75 & 87.05 {\scriptsize$\pm$} 0.66 & 82.80 {\scriptsize$\pm$} 2.74 & 57.32 {\scriptsize$\pm$} 0.20 & 55.29 {\scriptsize$\pm$} 0.12 & 0.499 {\scriptsize$\pm$} 0.001 & 0.806 {\scriptsize$\pm$} 0.004 & 109.7 {\scriptsize$\pm$} 0.0 & 50.5 {\scriptsize$\pm$} 0.4 \\
RoBERTa-L + data2vec + Facet $\rho{=}0.5$ & 3 & 87.21 {\scriptsize$\pm$} 0.79 & 87.20 {\scriptsize$\pm$} 0.69 & 82.94 {\scriptsize$\pm$} 2.91 & 57.37 {\scriptsize$\pm$} 0.14 & 55.27 {\scriptsize$\pm$} 0.10 & 0.498 {\scriptsize$\pm$} 0.001 & 0.807 {\scriptsize$\pm$} 0.004 & 105.6 {\scriptsize$\pm$} 0.0 & 47.9 {\scriptsize$\pm$} 0.8 \\
RoBERTa-L + data2vec + Facet $\rho{=}0.4$ & 3 & 86.95 {\scriptsize$\pm$} 1.06 & 86.97 {\scriptsize$\pm$} 0.97 & 82.46 {\scriptsize$\pm$} 2.71 & 56.55 {\scriptsize$\pm$} 0.18 & 54.59 {\scriptsize$\pm$} 0.24 & 0.505 {\scriptsize$\pm$} 0.005 & 0.804 {\scriptsize$\pm$} 0.004 & 94.9 {\scriptsize$\pm$} 0.0 & 40.1 {\scriptsize$\pm$} 2.4 \\
\bottomrule
\end{tabular}}
\end{table*}

\begin{table*}[t]
\centering
\caption{\textbf{Number of exits per modality $J$.} Exits are evenly spaced in each encoder. Each cell: Acc-2 / MAE (measured GFLOPs). ``--'': no configuration fits the budget, since the depth-only model cannot go below $1/J$ of the modelled cost.}
\label{tab:supp_exits_J}
\fittable{\textwidth}{%
\begin{tabular}{llccccc}
\toprule
Model & $J$ & full depth & $\rho{=}0.8$ & $\rho{=}0.6$ & $\rho{=}0.4$ & $\rho{=}0.25$ \\
\midrule
\multicolumn{7}{c}{\dsmosi{}} \\
\midrule
\method{} & 2 & 84.60 / 0.852 (139.2) & 85.37 / 0.896 (104.0) & 83.38 / 0.957 (91.3) & -- & -- \\
 & 3 & 90.09 / 0.605 (139.2) & 88.87 / 0.619 (92.3) & 89.02 / 0.640 (91.7) & 56.86 / 1.457 (74.9) & -- \\
 & 4 & 88.62 / 0.606 (139.2) & 88.77 / 0.619 (92.3) & 88.67 / 0.622 (86.9) & 75.41 / 0.938 (76.9) & 49.59 / 1.558 (66.2) \\
 & 6 & 89.94 / 0.574 (139.2) & 89.18 / 0.586 (80.6) & 89.18 / 0.586 (80.6) & 86.28 / 0.692 (77.4) & 53.20 / 1.495 (62.8) \\
 & 12 & 88.41 / 0.625 (139.2) & 88.57 / 0.639 (92.3) & 88.57 / 0.638 (91.8) & 87.80 / 0.645 (74.0) & 85.37 / 0.789 (64.6) \\
\addlinespace
\method-skip & 2 & 88.87 / 0.574 (139.2) & 89.63 / 0.567 (27.0) & 89.63 / 0.567 (27.0) & 89.63 / 0.567 (26.9) & 89.33 / 0.594 (25.0) \\
\bottomrule
\end{tabular}}
\end{table*}

\begin{table*}[t]
\centering
\caption{\textbf{Examples from the \dsmosi{} test set.} Each cell: predicted sentiment (selected text/audio/visual exit; ``--'': skipped). Labels and predictions are in $[-3,3]$; transcripts are those of the dataset.}
\label{tab:supp_cases}
\fittable{\textwidth}{%
\begin{tabular}{lccccccc}
\toprule
 & & \multicolumn{4}{c}{\method-skip} & \multicolumn{2}{c}{\method{}} \\
\cmidrule(lr){3-6}\cmidrule(lr){7-8}
Transcript & Label & full depth & $\rho{=}0.6$ & $\rho{=}0.4$ & $\rho{=}0.25$ & $\rho{=}0.4$ & $\rho{=}0.25$ \\
\midrule
\multicolumn{8}{l}{\emph{Text is sufficient}} \\
``but he wasn't even interesting'' & $-2.0$ & $-2.09$ & $-1.92$ (4/--/4) & $-1.92$ (4/--/4) & $-1.92$ (4/--/4) & $+1.01$ (2/1/2) & $+1.19$ (1/1/1) \\
``yeah it really is good I mean'' & $+2.2$ & $+2.28$ & $+2.05$ (4/--/4) & $+2.05$ (4/--/4) & $+2.05$ (4/--/4) & $-0.57$ (2/1/2) & $-0.48$ (1/1/1) \\
``I think um the movies are phenomenal especially that first one'' & $+2.8$ & $+2.54$ & $+2.48$ (4/--/4) & $+2.48$ (4/--/4) & $+2.48$ (4/--/4) & $+2.11$ (3/1/2) & $-0.52$ (1/1/1) \\
\addlinespace
\multicolumn{8}{l}{\emph{Skipping audio changes the prediction}} \\
``because I really wanted to see it actually'' & $+0.5$ & $+0.37$ & $+0.40$ (4/1/4) & $-0.05$ (4/--/4) & $-0.05$ (4/--/4) & $-0.12$ (4/1/4) & $+1.37$ (1/1/1) \\
``uh the dialogue is terrible'' & $-2.8$ & $-2.52$ & $-2.10$ (4/--/4) & $-2.10$ (4/--/4) & $-2.10$ (4/--/4) & $-2.16$ (3/1/2) & $-0.51$ (1/1/1) \\
``and I really really loved it'' & $+2.0$ & $+3.11$ & $+2.63$ (4/1/4) & $+2.38$ (4/--/4) & $+2.38$ (4/--/4) & $+2.36$ (4/1/4) & $+0.89$ (1/1/1) \\
\addlinespace
\multicolumn{8}{l}{\emph{Failure at a tight budget}} \\
``I could barely stay awake through the movie'' & $-2.6$ & $-1.98$ & $-1.95$ (4/--/4) & $-1.95$ (4/--/4) & $+0.30$ (1/--/1) & $+0.95$ (1/2/4) & $+1.15$ (1/1/1) \\
\bottomrule
\end{tabular}}
\end{table*}

\end{document}